\documentclass{article}
\usepackage{microtype}
\usepackage{graphicx}
\usepackage{subfigure}
\usepackage{booktabs} %
\usepackage{hyperref}

\usepackage[accepted]{icml2023}

\usepackage{amsmath}
\usepackage{amssymb}
\usepackage{mathtools}
\usepackage{amsthm}
\usepackage{relsize}
\usepackage{graphicx}
\usepackage{graphics}
\usepackage{caption}
\usepackage{subfigure}
\usepackage{amsmath}
\usepackage{booktabs}
\usepackage{multirow}
\usepackage{multicol}
\usepackage{mdframed}
\usepackage{color}
\usepackage{algorithm}
\usepackage{caption}
\usepackage[capitalize,noabbrev]{cleveref}
\theoremstyle{plain}

\theoremstyle{definition}

\theoremstyle{remark}

\usepackage[textsize=tiny]{todonotes}

\icmltitlerunning{Deep Regression Unlearning}

\begin{document}

\twocolumn[
\icmltitle{Deep Regression Unlearning}



\icmlsetsymbol{equal}{*}

\begin{icmlauthorlist}
\icmlauthor{Ayush K Tarun }{equal,yyy}
\icmlauthor{Vikram S Chundawat}{equal,yyy}
\icmlauthor{Murari Mandal}{comp}
\icmlauthor{Mohan Kankanhalli}{sch}
\end{icmlauthorlist}

\icmlaffiliation{yyy}{Mavvex Labs, India}
\icmlaffiliation{comp}{School of Computer Engineering, Kalinga Institute of Industrial Technology Bhubaneswar, India}
\icmlaffiliation{sch}{School of Computing, National University of Singapore}

\icmlcorrespondingauthor{Murari Mandal}{murari.nus@gmail.com}
\icmlkeywords{Machine Unlearning, Regression Unlearning, Data Privacy, Deep Learning}
\vskip 0.3in
]



\printAffiliationsAndNotice{\icmlEqualContribution} 
\begin{abstract}
With the introduction of data protection and privacy regulations, it has become crucial to remove the lineage of data on demand from a machine learning (ML) model. In the last few years, there have been notable developments in machine unlearning to remove the information of certain training data efficiently and effectively from ML models. In this work, we explore unlearning for the regression problem, particularly in deep learning models. Unlearning in classification and simple linear regression has been considerably investigated. However, unlearning in deep regression models largely remains an untouched problem till now. In this work, we introduce deep regression unlearning methods that generalize well and are robust to privacy attacks. We propose the \emph{Blindspot} unlearning method which uses a novel weight optimization process. A randomly initialized model, partially exposed to the retain samples and a copy of the original model are used together to selectively imprint knowledge about the data that we wish to keep and scrub off the information of the data we wish to forget. We also propose a Gaussian fine tuning method for regression unlearning. The existing unlearning metrics for classification are not directly applicable to regression unlearning. Therefore, we adapt these metrics for the regression setting. We conduct regression unlearning experiments for computer vision, natural language processing and forecasting applications. Our methods show excellent performance for all these datasets across all the metrics. Source code: \url{https://github.com/ayu987/deep-regression-unlearning}
\end{abstract}



\section{Introduction}
\label{sec:intro}

Data is an essential asset of any organization and it has opened up a new frontier for countries to flex their technological and economic muscle. Governments across the globe have taken cognizance of the importance of data privacy and protection. The data protection law, European Union General Data Protection Regulation (EU GDPR)~\cite{voigt2017eu}, introduced in the European Union has changed the way companies handle personal data. Similarly, in USA, the California Consumer Privacy Act (CCPA)~\cite{goldman2020introduction} has been introduced in California to protect the privacy of users and give them more control over the use of their data. The introduction of these rules have engendered a set of changes in the way organizations collect, store, analyze, and use personal data collected from citizens. In particular, all users are given the \textit{right to be forgotten} under these data protection regulations. The EU GDPR necessitates prior consent by the user to collect their data. The CCPA allows the company to collect user data by default. However, the user may request for removal of his/her data at any point in time. A company is obligated to remove the data pertaining to a user upon receiving a request for deletion.\par

In case of simple aggregation and storage of data, it is easy to remove the data from the company's databases. However, a machine learning (ML) model trained on such personal data essentially creates a new type of data which is an indirect representation of the original data. The enforcement of the \textit{right to be forgotten} on such ML models will help control the analytical use of data through ML algorithms that  do not strictly fall under the purview of traditional understanding of data privacy. However, the removal of indirectly represented information from ML models is a non-trivial problem. A typical machine learning algorithm learns about the data by observing a large number of data samples. The information about the data is encoded in the weights of the ML model. This means the model weights contain information about the data. Any request for removal of information about a particular data or set of data would require manipulating the set of weights in the model. In a general ML setting, the model is trained using the training dataset. After the model is optimized via some learning method, it is used for inference in the downstream application. Upon receipt of a data removal request, the information pertaining to the \textit{forget data} is required to be scrubbed from the model. A naïve approach is to retrain the model from scratch after excluding the data that needs to be removed/forgotten. However, this is not a feasible solution due to limited resources available to repeatedly train the ML model. For large models, this approach would make the response time very high which might not be acceptable to the user or the compliance authority. Besides, the unnecessary use of energy-intensive GPU servers would add to the already acute problem of climate change. An efficient approach would be to update the model weights in such a way that the information is forgotten by the model. An unlearning method should ideally provide an effective, efficient, and robust solution to enable such a change in the model.\par

\textbf{Motivation.} Machine unlearning is an important field of study as the existence of effective and efficient machine unlearning solutions would give confidence to the lawmakers to formulate stricter data privacy and protection policies for their citizen. The existing works in unlearning have primarily focused on the classification problems. Relatively simple models such as linear and logistic regression~\cite{mahadevan2021certifiable, neel2021descent,izzo2021approximate,guo2020certified}, random forests~\cite{brophy2021machine}, and k-means clustering~\cite{ginart2019making,mirzasoleiman2017deletion} have been explored in an provable unlearning setup. Furthermore, the deep learning models such as convolutional neural networks~\cite{golatkar2020eternal,golatkar2020forgetting,golatkar2021mixed,wang2022federated,wu2022federated,chundawat2022zero} and vision transformers~\cite{tarun2021fast} have been explored under the approximate unlearning setup. All these existing methods are aimed at unlearning in classification problems. Li et al.~\cite{pmlr-v130-li21a} proposed an online forgetting process for linear regression models. This method does not generalize to deep learning regression models. Unlearning in a regression problem, particularly if it employs a deep learning approach, is yet to be explored in the literature. Same is the case with unlearning in case of forecasting models. The existing methods designed for classification tasks cannot be directly applied to these tasks. Moreover, the evaluation metrics for unlearning in a classification task do not transfer well to unlearning in a regression task.\par


\textbf{Our Contribution.} Based on the above motivation, in this paper, we propose novel unlearning methods for deep regression models. The proposed \emph{Blindspot} method selectively removes the information of the forget data and keeps the information of the retain data through a collaborative optimization process. We use a randomly initialized model and partially expose it to the retain samples. Then another model, initialized with the original model's weights is optimized with three loss functions: i) the attention difference loss, ii) forget data output difference loss between the partially exposed model and the original model, and iii) retain data prediction loss which helps in maintaining the original prediction accuracy on the retain data. In effect, the relevant knowledge of the original model is selectively imprinted into the new model and unlearning of the forget samples is induced through our novel weight optimization process. Our strategy ensures that the unlearning occurs both at the representation level and the ingrained level. We also present a Gaussian distribution based fine tuning method for regression unlearning. We check for privacy leaks in the unlearned model by designing a membership inference attack and  inversion attack for regression problem. Several ablation studies are conducted to show the characteristics of the proposed method. In summary, the main contributions of our paper are:

\begin{enumerate}
    \item \textbf{Novelty:} To the best of our knowledge, this work is the first to study unlearning in deep regression models and forecasting. We propose two deep regression unlearning methods that achieve quality unlearning with good performance and is robust to privacy attacks. 
    
    \item \textbf{Unlearning in Deep Regression Models:} We propose a Blindspot method which optimizes three loss functions to induce selective unlearning. We also propose a Gaussian Amnesiac learning method for regression unlearning.

    \item \textbf{Effectiveness:} We conduct extensive experiments on four datasets AgeDB, IMDB-Wiki, STS-B (SemEval-2017) and UCI Electricity load. The results show that the proposed method outperforms the baseline methods for regression unlearning on a variety of metrics that denote both representation and ingrained level unlearning.

    \item \textbf{Robustness to Privacy Attacks:} The proposed unlearning methods are resistant to membership inference and inversion attacks. This provides more confidence to the user regarding  privacy preservation  against queries related to his/her forget data.
\end{enumerate}
\section{Related Work}
\label{sec:related}
Machine unlearning has been investigated for different tasks and different modalities of learning algorithms. Generally, these methods can be categorized into exact and approximate unlearning techniques. Exact unlearning aims to completely remove the requested data from the model. Approximate unlearning aims to provide a statistical guarantee that the unlearned model cannot be distinguished from a model that was trained without using the forget data. The unlearning can be measured at both the representation (abstract level) and ingrained level (model weight level information) in order to offer the statistical guarantees. We discuss the existing methods in the literature as designed for the classification and regression tasks.

\textbf{Unlearning in Classification Tasks.} Cao et al.~\cite{cao2015towards} introduced machine unlearning to selectively remove the effect of a subset of training data. Several research subsequently aimed to produce efficient and effective ways of unlearning. In SISA framework~\cite{bourtoule2021machine} the model learns from the summation of different subsets of data. Chen et al.~\cite{chen2022graphunlearning} extend this idea to unlearning in graph data. An unlearning framework for recommendation is presented in~\cite{chen2022recommendation}. In Amnesiac learning~\cite{graves2021amnesiac} the updates made by each data-point is stored during training and subtracted from final parameters upon each deletion request. The definition of differential privacy is adopted to introduce a probabilistic notion of unlearning in~\cite{ginart2019making}. The idea is to produce similar distribution of output between the unlearned model and the model trained without using the forget data. This approach is frequently used in several methods~\cite{mirzasoleiman2017deletion,izzo2021approximate,ullah2021machine}. A certified data removal framework is presented in~\cite{guo2020certified}.~\cite{neel2021descent} use gradient descent based approach for unlearning in convex models. Unlearning for Bayesian methods~\cite{nguyen2020variational}, k-means clustering~\cite{mirzasoleiman2017deletion}, random forests~\cite{brophy2021machine} and other studies~\cite{sekhari2021remember,warnecke2021machine,mahadevan2021certifiable} have been explored.\par

Some of the early works on unlearning in convolutional neural networks (CNN) were presented in~\cite{golatkar2020eternal}. This work presents a scrubbing method to remove information from the network weights. A neural tangent kernel (NTK) based method to approximate the training process was introduced in~\cite{golatkar2020forgetting}. An approximated model is used to estimate the network weights for the unlearned model. Similarly, a mixed-linear model is trained for unlearning approximation in~\cite{golatkar2021mixed}. A more practical and efficient approach for deeper neural networks and vision transformers was presented in~\cite{tarun2021fast}. A zero-shot unlearning method was presented in~\cite{chundawat2022zero}. A teacher-student based framework for class-level unlearning as well as random cohort unlearning was introduced in~\cite{chundawat2022can}. Other notable works in deep unlearning include~\cite{mehta2022deep,ye2022eccvlearning}. Several works have presented efficient methods for unlearning in the federated learning setup~\cite{wang2022federated,liuinfocomright2022,liu2021revfrf}.~\cite{pmlr-v162-bevan22a} use the bias unlearning methods~\cite{kim2019learning} to remove bias from CNN based melanoma classification. Some recent works have identified the vulnerabilities of the unlearned model under different type of attacks~\cite{marchant2022hard,carlini2022privacy,chen2021machine}.

\textbf{Unlearning in Deep Regression Tasks.}~\cite{pmlr-v130-li21a} investigated online forgetting process in ordinary linear regression tasks. The method supports a class of deletion practice \textit{first in first delete} where the user authorize the use of their data for limited period of time.~\cite{izzo2021approximate} proposed an approximate deletion method for linear and logistic regression. The existing methods are relevant for convex models and are hard to apply on non-convex models like deep neural networks for regression unlearning. Our method does not put any constraint over the underlying model used. Similarly, our method does not require prior information related to the training procedure as in some existing works~\cite{nguyen2022survey}. From the survey paper~\cite{nguyen2022survey}, it is evident that there is no existing work on deep regression unlearning and ours is the first deep regression unlearning method. This work introduces the first deep regression unlearning methods for deeper models and large datasets. Our work also presents a set of suitable metrics for evaluation of the unlearned regression models. 


\section{Regression Unlearning}
\subsection{Preliminaries}
Let $D = \{x_i, y_i\}_{i=1}^N$ be a dataset consisting of $N$ samples where $x_i \in \mathbb{R}$ is the $i^{th}$ sample, and $y_i \in \mathbb{R}$ is the corresponding output variable. $D_f$ denotes the set of data-points we wish to forget. These are the data-points a machine unlearning algorithm will receive as a query. They may or may not be related in any way. Similarly, $D_r$ denotes the set of data-points whose knowledge we wish the model to retain. This means $D = D_r \cup D_f$, $D_r$ and $D_f$ are mutually exclusive i.e., $D_r \cap D_f = \phi$. The model trained from scratch with only $D_r$ is called the \textit{retrained} model or \textit{gold} model in this work.\par

To measure the similarity between output distributions of different models, we use the first Wasserstein Distance~\cite{kantorovich1960mathematical,ramdas2017wasserstein}. We treat the output space as a metric. Let $p$ be the output distribution of model 1 and $q$ be the output distribution of model 2, then the first Wasserstein Distance between these two distributions is defined by
\begin{equation}
\label{eq:wasserstein}
    {W_{1}(p ,q ) = \inf _{\gamma \in \Gamma (p ,q )}\int _{\mathbb{R}\times \mathbb{R}}|x-y|\,\mathrm {d} \gamma (x,y)}
\end{equation}
where $\Gamma (p ,q )$ is the set of probability distributions on $\mathbb{R}\times\mathbb{R}$  whose marginals are $p$ and $q$ on the first and second factors respectively.

\subsection{Problem Formulation}
Let $M(.;\phi)$ be a machine learning (ML) model $M$ with parameters $\phi$. For an input $x$ the model returns $M(x;\phi)$. For a ML algorithm $A$ trained on dataset $D$, the obtained model parameters $\phi$ can be represented as
\begin{equation}
    \phi = A(D)
\end{equation}
The parameters of a model trained only on the retain set of $D$ i.e., $D_r$ (also called retrained model) is represented as 
\begin{equation}
    \phi_r = A(D_r)
\end{equation}
A machine unlearning algorithm $U$ uses the originally trained model $\phi$. It may also use a subset of $D_r$ and $D_f$. With this algorithm $U$ we obtain a new set of parameters $\phi_u$ as follows.
\begin{equation}
    \phi \underset{U(\phi,D_r,D_f)}{\to} \phi_u
\end{equation}
An exact unlearning algorithm aims to produce a model with exactly the same output distribution as that of the retrained model. In this work, we propose an approximate unlearning algorithm which aims to obtain a parameter set $\phi_u$ which results in approximately the same output distribution as that of the retrained model i.e.,
\begin{equation}\label{eq:similarity}
    P(M(x, \phi_u)=y) \approx P(M(x, \phi_r)=y)  \; \forall x \in D, y \in \mathbb{R}
\end{equation}
where $P(X)$ denotes the probability distribution of any random variable $X$. Note that we emphasize only on the similarity between the output distributions and not the parameters. Readout functions~\cite{golatkar2020eternal} are used to check the validity of Eq.~\ref{eq:similarity} and the validity/quality of $\phi_u$ obtained using $U$.

\begin{figure}[]
\centering
    \includegraphics[width=0.5\textwidth]{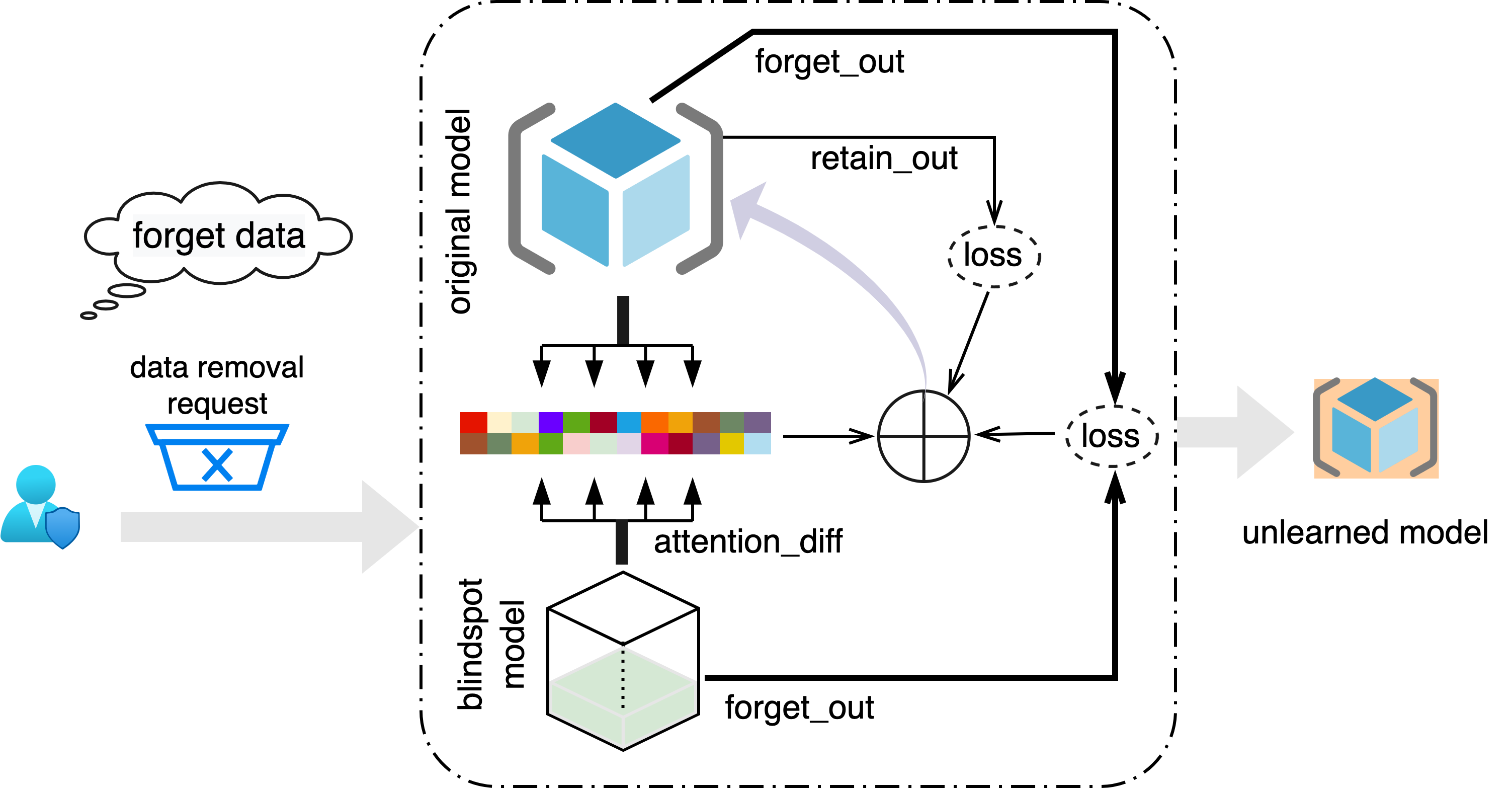}
\caption{The proposed Blindspot deep regression unlearning method. The \textit{blindspot model} is first partially trained for a few epochs on the retain set and frozen during the unlearning process.}
\label{fig:proposed_framework}
\end{figure}

\subsection{Challenges in Regression Unlearning}
The existing machine unlearning methods use measures like KL-Divergence with a retrained model, or a proxy to the retrained model, or provide certain information bounds for the unlearned model. The KL-Divergence~\cite{golatkar2020eternal} is used as a measure of closeness between the current and the desired distribution. Unlike in classification, where we have a probability distribution, in regression we usually deal with a single valued output. In a regression task, we generally predict the expected value of the real-valued label instead of a set of probabilities associated with each label. So, it becomes non-trivial to ascertain the output distribution based on the obtained output. Thus, regression unlearning is different and a difficult task. Moreover, the inference attacks in a regression setting is not well studied in the literature leading to another challenge in evaluation of regression unlearning.

\section{Proposed Deep Regression Unlearning}
We propose two algorithms to delete information about the query data from a deep regression model: (i) Blindspot Unlearning, (ii) Gaussian Amnesiac Learning. We discuss each of these methods below. 

\subsection{Blindspot Unlearning}
\label{sec:blindspot}
In Blindspot unlearning, we first partially expose a randomly initialized model to few samples from the retain set. It is trained on the retain samples for a few epochs. This gives the model a vague idea about the output distribution in the absence of the forget set from the training data. 
The forget set is a \textit{blindspot} for this model. This partially learned blindspot model acts as an \textit{unlearning helper}. Let the blindspot model be denoted as $B(.;\theta)$. We denote the original fully trained model by $M(x_{i};\phi)$. In our method, the model $M$ is updated to obtain the final unlearned model. The complete Blindspot Unlearning method is depicted in Figure~\ref{fig:proposed_framework}. The following three loss functions are employed in our method: (i) We compute the loss for the retain set sample prediction in the original model, (ii) We compute the loss by comparing the output similarities between the original and the blindspot model, (iii) We also measure the closeness of the layerwise activation between the original and blindspot model. We combine these three losses and optimize the original model through minimization. This process selectively keeps the knowledge regarding the retain set while removing the knowledge about the forget set. Let the prediction made by the original model on $i^{th}$ sample of dataset $D$ is $M(x_{i};\phi)$ and $y_i$ is the corresponding correct label. Then the loss for samples in $D_r$ is
\begin{equation}
    L_{r} \gets \mathcal{L}(M(x_{i};\phi), y_i); \forall {x_{i}\in D_{r}}
\end{equation}
where $\mathcal{L}$ denotes a standard loss function used in a regression task. This can be a mean absolute error (MAE), mean squared error (MSE), or some other regression loss function. Let $M(x_{i};\phi)$ denote the prediction of fully trained model on sample $x_i$ of dataset $D$. Similarly, let $B(x_{i};\theta)$ denote the prediction of the blindspot model. If the sample $x_i$ is a part of the forget set $D_f$, then the following loss is computed
\begin{equation}
    L_{f} \gets \mathcal{L}(M(x_{i};\phi), B(x_{i};\theta)); \forall {x_{i}\in D_{f}}
\end{equation}
This loss helps in increasing the closeness between the original model $M$ and the blindspot model $B$ for the forget set $D_f$. Since we want to scrub the information related to the forget data from the unlearned model. The blindspot model $B$ works as the perfect foil for it to unlearn the same. Finally, we optimize the closeness of activations~\cite{micaelli2019zero} between the last $k$ layers of model $M$ and $B$ on the forget set $D_f$ 
\begin{equation}
    L_{attn} \gets \lambda\sum_{j=1}^{k} \|act^{\phi}_{j}-act^{\theta}_{j}\|
\end{equation}
where $act^{\phi}_{j}$ and $act^{\theta}_{j}$ corresponds to the $j^{th}$ layer of activation map in the original model $M$ and blindspot model $B$. $\lambda$ is a parameter used to control the relative degree of significance of the loss terms. The final loss is computed as
\begin{equation} \label{eq:blindspot}
    L \gets (1-l^i_f)L_{r} + l^i_f(L_{f} + L_{attn})
\end{equation}
where $l^i_f=1$ for samples in the forget set and  $l^i_f=0$ otherwise. A step-by-step process of the proposed Blindspot Unlearning method is given in Algorithm~\ref{alg:blindspot}. The information present in the unlearned model about the forget set after unlearning is bounded by the information present in the blindspot model. More details on the information bound for unlearning in the Blindspot method is discussed in Section~\ref{sec_inf_bound} in the Supplementary. 

\begin{algorithm}[t]
\caption{Blindspot Unlearning}
\label{alg:blindspot}
\begin{algorithmic}[1]
\STATE M(.;$\phi$) (Fully Trained Model)
\STATE B(.;$ \theta$) (Randomly Initialized Blind model)
\STATE $D_f \gets$ forget set (from training data)
\STATE $D_r \gets$ retain set (from training data)
\STATE $D = D_r \cup D_f$
\FOR{1,2....$n$}
\STATE Partially expose model to retain samples with very less epochs $n<<<n_{epochs}$ 
\FOR{$x_{i},y_{i} \in D_r$}
\STATE $y^{pred}_{i}$ $\gets$ $B(x_{i};\theta)$
\STATE $L \gets \mathcal{L}(y^{pred}_{i}, y_{i})$
\STATE $\theta \gets \theta - \eta \frac{\partial L}{\partial \theta}$, where $\eta$ is the learning rate
\ENDFOR
\ENDFOR
\FOR{1,2....$n_{unlearn}$}
    \FOR{$(x_{i},y_{i}) \in D$} 
    \STATE $y^{pred}_{i}$ $\gets$ $M(x_{i};\phi)$ (Finetune)

    \IF{$(x_{i},y_{i}) \in D_f$}
        \STATE $l^i_f$ = 1 (Forget Label)
    \ELSE
        \STATE $l^i_f$ = 0
    \ENDIF
    
    \STATE $L_r \gets \mathcal{L}(y^{pred}_{i}, y_{i})$
    \STATE $L_f \gets \mathcal{L}(y^{pred}_{i}, B(x_{i};\theta))$
    \STATE $L_{attn} \gets \lambda\sum_{j=1}^{k} \|act^{\phi}_{j}-act^{\theta}_{j}\|$
    \STATE $L \gets (1-l^i_f)L_r + l^i_f(L_f + L_{attn})$
    \STATE $\phi \gets \phi - \eta \frac{\partial L}{\partial \phi}$
    \ENDFOR
\ENDFOR
\end{algorithmic}
\end{algorithm}

\subsection{Gaussian-Amnesiac Learning} 
\label{sec:reg-amnesiac}
We adapt the unlearning technique in~\cite{graves2021amnesiac} which was originally presented for a classification task. In this method, the label of a sensitive data is replaced with an incorrect label. In a classification problem, it is reasonable to assume that the samples are uniformly distributed across class labels. However, this is almost never the case in a regression problem and real-life regression data usually resemble a Gaussian distribution~\cite{bishop2006pattern}. Thus, in a regression task, we model the distribution of the regression output values as a Gaussian model. The incorrect labels are sampled from this Gaussian distribution instead of random assignment. A straightforward adaptation by replacing random selection with a uniform distribution produces inferior results. This is shown through experiments in Section~\ref{sec:ablation_gaussian_vs_regular_Amnesiac} where we compare the results of sampling from a Gaussian distribution with a uniform distribution. We then fine tune the original model on this data. Algorithm~\ref{alg:reg-amnesiac} shows the step-by-step process of the Gaussian Amnesiac learning.

\begin{algorithm}[t]
\caption{Gaussian-Amnesiac Learning}
\label{alg:reg-amnesiac}
\begin{algorithmic}[1]
\STATE M(.;$\phi$) (Fully Trained Model)
\STATE $D_f \gets$ forget set (from training data)
\STATE $D_r \gets$ retain set (from training data)
\STATE $D^{'}_{f} \gets [\;]$
\FOR{$(x_{i},y_{i}) \in D_f$}
\STATE Replace labels of forget samples: $y^{'}_{f} \gets \mathcal{N(\mu, \sigma)}$ (Sample random labels from a Gaussian distribution, $\mathcal{N(\nu, \sigma)}$ of labels $Y=\{y_{i} \forall (x_{i},y_{i}) \in D_{f}\}$)
\STATE $D^{'}_{f}$ = $D^{'}_{f}$ + $(x_{i},y^{'}_{i})$
\ENDFOR
\STATE $D^{'}$ $\gets$ $D_r$ + $D^{'}_{f}$ (New dataset to finetune the original model)
\STATE shuffle($D^{'}$)
\FOR{1,2....$n_{finetune}$}
\FOR{$(x_{i},y_{i}) \in D^{'}$}
\STATE $y^{pred}_{i}$ $\gets$ $M(x_{i};\phi)$ (Finetune)
\STATE $L_M \gets \mathcal{L}$($y^{pred}_{i}$, $y_{i}$)
\STATE $\phi \gets \phi - \eta \frac{\partial L_M}{\partial \phi}$
\ENDFOR
\ENDFOR
\end{algorithmic}
\end{algorithm}

\section{Evaluation Measures}
A machine unlearning method is evaluated through a variety of measures in the literature~\cite{golatkar2021mixed,chundawat2022can}. These metrics usually validate the unlearning at the representation level and ingrained level. At representation level, the unlearning is validated through the \textit{model error or accuracy} on the forget set and retain set. The \textit{relearn time} to achieve similar performance as the original model~\cite{golatkar2020eternal,chundawat2022zero} also falls into this category. The ingrained level evaluation include the weight and output distribution analysis of the unlearned model. Several metrics such as activation distance, weight distance, JS-Divergence, ZRF-score~\cite{chundawat2022can} comes under this category. Prediction distribution analysis on the forget class~\cite{tarun2021fast} is another type of ingrained level evaluation. A third class of evaluation entails checking the privacy leakage about the forget data in the unlearned model through various types of inference attacks. In our work, \textit{we validate the deep regression unlearning methods with all three categories of evaluation methods.}

\textbf{Privacy Attacks: Membership Inference and Model Inversion Attacks for Regression}
We develop a simple membership inference attack to evaluate the regression unlearning approaches in this study. We construct the membership attack as a binary classification problem where class 1 denotes a data point is in the training set and 0 denotes it is in the test set. We use a support vector classifier with radial basis function kernel as the attacker. We use \textit{loss, penultimate layer gradients, and penultimate layer activations} as the inputs to the attacker for classification. We train the classifier by providing retain set as class 1 and test set as class 0. We use this trained attacker on the forget set.\par 
For inversion attack, we use a modified version of the attack presented in~\cite{fredrikson2015model}. A randomly initialized image vector is optimized using gradient descent using mean squared error as the loss function. Section~\ref{sec:inversion} in the Supplementary shows the model inversion attack results.\par

\textbf{Relearning Effort: Regression Anamnesis Index.} We adapt the Anamnesis Index (AIN) proposed in~\cite{chundawat2022zero} to measure the relearning effort in the unlearned regression model. The AIN measures the relearning time of the retrained model (retrained from scratch without forget data) and unlearned model to come under $\alpha\%$ margin of the performance of a fully trained model. Let $M_u$ and $M_g$ denote the unlearned model and the retrained model on $D_r$, respectively. If the number of mini-batches (steps) required by a model $M$ to come within $\alpha\%$ range of the accuracy of the original model on the forget classes is $rt(M, M_{orig}, \alpha)$ then
\begin{equation}
AIN = \frac{rt(M_u, M_{orig}, \alpha)}{rt(M_g, M_{orig}, \alpha)}
\end{equation}
For our regression use case, we define $rt(M, M_{orig}, \alpha)$ as the number of steps required to come within $\alpha\%$ range of the loss of the original model $M_{orig}$ on forget set. As discussed in~\cite{chundawat2022zero}, AIN close to 0 denotes sub-optimal unlearning and AIN close to 1 denotes an adequate amount of unlearning. If the AIN is very large, then it signifies the \textit{Streisand effect} where the sample to be forgotten is actually made more noticeable.\par

\textbf{Output Distribution: Wasserstein Distance.} In case of class-level unlearning, KL-Divergence and JS-Divergence~\cite{golatkar2020eternal,chundawat2022can} between the output distribution of retrained and the unlearned model is compared. In our analysis, we use~\textit{Wasserstein Distance} (refer Eq.~\ref{eq:wasserstein}) between the forget set prediction of the retrained and the unlearned model. We also measure the \textit{relative deviation} for each individual prediction on the forget set and plot the density curves for the same. If the density is closer to zero, then the unlearning is better in the model.

\textbf{Performance: Error on $D_f$ and $D_r$.} Unlike in a classification task, the metrics in a regression task are usually not bounded (for example, mean squared error, mean absolute error). Therefore these measures are not fit for checking the quality of unlearning in a regression model. In our work, we report the error on both forget and retain set. The errors should be close to the corresponding metrics on the retrain model.
\begin{table*}[]
\tiny
\footnotesize
\centering
\caption{Unlearning on AgeDB. We unlearn samples from a specific age band and observe the performance on several unlearning metrics.~$err\_D^{r}_{t}$: error on retain set from test data, $err\_D^{f}_{t}$: error on forget set from train data, $att\_prob$: membership inference attack probability on forget set, $w\_dist$: Wasserstein distance between the unlearned and retrained model predictions on $D^{f}_{t}$, $AIN$: Anamnesis Index, \textit{Amn}: Amnesiac. A ResNet18 model is used in all the experiments.}
\begin{tabular}{c|c|cccccc}
\hline
Forget Set &  Metric & Original & Retrained& FineTune & NegGrad & \textbf{Gaussian Amn (Ours)} &\textbf{Blindspot (Ours)}\\
\hline
\multirow{5}{*}{0-30} &  $err\_{D_t^r}$ $\downarrow$ & 7.69 & 7.54 & 7.59 $\pm$ 0.32 & 23.01 $\pm$ 1.12 & {7.51 $\pm$ 0.19} & 7.63 $\pm$ 0.27\\
{} & $err\_{D_t^f}$ $\uparrow$ & 8.11 & 15.1 & 10.40 $\pm$ 0.28 & 30.47 $\pm$ 0.31 & 13.73 $\pm$ 0.22 & {18.27 $\pm$ 0.24}\\
{} & {$att\_prob \downarrow$} & {0.72} & {0.07} & {0.51 $\pm$ 0.03} & \textbf{0 $\pm$ 0} & {0.13 $\pm$ 0.01} & {{0.02 $\pm$ 0}}\\
{} & $w\_dist \downarrow$ & {11.39} & {-} & 7.40 $\pm$ 0.13 & 12.82 $\pm$ 0.17 & 3.74 $\pm$ 0.09 & \textbf{1.90 $\pm$ 0.06}\\
{} & AIN $\uparrow$ & {-} & {-} & 1.6 $\pm$ 0.20 & 0.33 $\pm$ 0.04 & 0.66 $\pm$ 0.03 & \textbf{1 $\pm$ 0.04}\\

\hline
 \multirow{5}{*}{60-100} & $err\_{D_t^r}$ $\downarrow$ & 7.24 & 6.73 & 6.78 $\pm$ 0.29 & 13.79 $\pm$ 0.20 & {6.73} $\pm$ 0.17 & 7.31 $\pm$ 0.18\\
{} & $err\_{D_t^f}$ $\uparrow$ & 10.43 & 22.87 & 13.5$\pm$0.24 & 34.87$\pm$ 0.97 & 21.01$\pm$ 0.64 & {25.8$\pm$ 0.53}\\
{} & {$att\_prob \downarrow$} & {0.62} & {0.03} &\textbf{0$\pm$ 0} & {0.10$\pm$ 0.02} & {0.02$\pm$ 0} & \textbf{0.01$\pm$ 0}\\
 {} & $w\_dist \downarrow$ & {17.74} & {-} & {11.81$\pm$ 0.41} & 11.34$\pm$ 0.34 & \textbf{2.71$\pm$ 0.23} & {3.66$\pm$ 0.21}\\
{} & AIN $\uparrow$ & {-} & {-} & 0.03 & \textbf{0.75$\pm$ 0.10} & 0.28$\pm$ 0.02 & {0.41$\pm$ 0.03}\\
\hline
\end{tabular}
\label{tab:agedb}
\end{table*}

\begin{table*}[t]
\tiny
\footnotesize
\centering
\caption{Unlearning on IMDBWiki. A ResNet18 model is used in all the experiments.}
\begin{tabular}{c|c|ccccc}
\hline
Forget Set & Metric & Original & Retrained & FineTune & \textbf{Gaussian Amn (Ours)} &\textbf{Blindspot (Ours)}\\
\hline
\multirow{5}{*}{0-30} & $err\_{D_t^r}$ $\downarrow$ & 8.27 & 7.52 & {7.86} & 7.94 & 8.04\\
{} & $err\_{D_t^f}$ $\uparrow$ & 7.64 & 17.34 & 14.15 & 16.12 & {20.66}\\
{} & {$att\_prob \downarrow$} & {0.75} & {0.13} & {0.26} & {0.14} & \textbf{0.07}\\
{} & $w\_dist \downarrow$ & {9.26} & {-} & {3.16} & \textbf{1.62} & {3.84}\\
{} & AIN $\uparrow$ & {-} & {-} & 0.005 & 0.01 & \textbf{0.04}\\
\hline
 \multirow{5}{*}{60-100} & $err\_{D_t^r}$ $\downarrow$ & 6.55 & 6.43 & {6.35} & 6.61 & 6.68\\
{} & $err\_{D_t^f}$ $\uparrow$ & 11.82 & 20.36 & 16.21 & 24.44 & 25.77\\
{} & {$att\_prob \downarrow$} & {0.56} & {0.06} & {0.33} & {0.0005} & \textbf{0.0}\\
{} & $w\_dist \downarrow$ & {10.68} & {-} & {5.55} & \textbf{4.03} & {5.05}\\
{} & AIN $\uparrow$ & {-} & {-} & 0.001 & {0.03} & \textbf{0.04}\\
\hline

\end{tabular}
\label{tab:imdb-wiki}
\end{table*}
\section{Experiments}
\subsection{Datasets} 
We use four datasets in our experiments. Two computer vision datasets are used: i. AgeDB~\cite{moschoglou2017agedb} contains 16,488 images of 568 subjects with age labels between 1 and 101, ii. IMDB-Wiki~\cite{rothe2015dex} contains 500k+ images with age labels varying from 1 to 100. One NLP dataset is used: iii. Semantic Text Similarity Benchmark (STS-B) SemEval-2017 dataset~\cite{cer2017semeval} has around 7200 sentence pairs and labels corresponding to the similarity between them on a scale of 0 to 5 categorized by genre and year. One forecasting dataset is used: iv. UCI Electricity Load dataset~\cite{yu2016temporal} contains data of electricity consumption of 370 customers, aggregated on an hourly level.

\begin{table*}[t]
\footnotesize
\centering
\caption{Unlearning results on Semantic Text Similarly Benchmark (STS-B) SemEval-2017 dataset.}
\begin{tabular}{c|c|ccccc}
\hline
Forget Set & Metric & Original & Retrained & FineTune & \textbf{Gaussian Amn (Ours)} &\textbf{Blindspot Ours)}\\
\hline
\multirow{4}{*}{0-2} & $err\_{D_t^r} \downarrow$ & 1.63 & 0.95 & 1.03 & 1.05 & {0.99}\\
{} & $err\_{D_t^f} \uparrow$ & 1.40 & 2.75 & 2.35 & 2.30 & {2.47}\\
{} & {$att\_prob \downarrow$} & {0.67} & 0.002 & 0.06 & 0.06 & \textbf{0.03}\\
{} & $w\_dist \downarrow$  & 1.64 & {-} & 0.64 & 0.63 & \textbf{0.35}\\
{} & AIN $\uparrow$ & {-} & {-} & 0.54 & 0.54 & \textbf{0.62}\\

\hline
{} & $err\_{D_t^r} \downarrow$ & 1.46 & 1.46 & 1.46 & 1.48 & 1.49\\
{Random} & $err\_{D_t^f} \uparrow$ & 1.35 & 1.49 & 1.34 & 1.35 & 1.41\\
{Samples} & {$att\_prob \downarrow$} & 0.73 & 0.60 & 0.54 & 0.62 & \textbf{0.53}\\
{1000} & $w\_dist \downarrow$  & 0.09 & {-} & 0.10 & 0.28 & \textbf{0.09}\\
{} & AIN $\uparrow$ & {-} & {-} & 0.03 & 0.03 & \textbf{1.0}\\

\hline
{} & $err\_{D_t^r} \downarrow$ & {1.46} & {1.46} & {1.46} & {1.47} & {1.48}\\
{Year} & $err\_{D_t^f} \uparrow$ & {1.49} & {1.77} & {1.52} & {1.57} & {1.62}\\
{2015} & {$att\_prob \downarrow$} & {0.70} & {0.52} & {0.50} & {0.47} & \textbf{0.34}\\
{Samples} & $w\_dist \downarrow$  & 0.32 & {-} & {0.11} & \textbf{0.03} & {0.05}\\
{} & AIN $\uparrow$ & {-} & {-} & 0.1 & 0.2 & \textbf{0.53}\\

\hline
\end{tabular}
\label{tab:sts-b}
\end{table*}

\begin{table*}[t]
\tiny
\footnotesize
\centering
\caption{Unlearning results on UCI Electricity Load dataset. \textit{Loss: the Quantile loss used in training}.}
\begin{tabular}{c|c|ccccc}
\hline
Forget Set& Metric & Original & Retrained & FineTune & \textbf{Gaussian Amn(Ours)} &\textbf{Blindspot (Ours)}\\
\hline
\multirow{5}{*}{$\leq-0.85$} & Loss on $D_t^r$ $\downarrow$ & 0.95 & 0.87 & 0.93 & 0.91 & 0.85\\
{} & Loss on $D_t^f$ $\uparrow$ & 0.87 & 0.90 & 1.40 & 1.40 & 1.25\\
{} & {$att\_prob \downarrow$} & 0.49 & 0.13 & 0.34 & 0.28 & \textbf{0.26}\\
{} & $w\_dist \downarrow$ & 15.43 & {-} & 0.54 & 2.25 & \textbf{0.13}\\
\hline
\multirow{5}{*}{$\geq0.85$} & Loss on $D_t^r$ $\downarrow$ & 0.82 & 0.61 & 0.71 & 0.77 & 0.74\\
{} & Loss on $D_t^f$ $\uparrow$ & 1.22 & 1.29 & 1.37 & 1.27 & 1.43\\
{} & {$att\_prob \downarrow$} & 0.23 & 0.20 & 0.29 & 0.36 & \textbf{0.17}\\
{} & $w\_dist \downarrow$ & 1.90 & {-} & \textbf{0.18} & 0.28 & 0.90\\
\hline
\end{tabular}
\label{tft-elec}
\end{table*}

\subsection{Baselines and Models}
We use fine tuning and gradient ascent (denoted as NegGrad) as baseline methods. In case of finetuning, the original model is fine tuned on the retain dataset $D_{r}$. The training only on the retain set leads to unlearning on the forget set $D_{f}$. This is same as catastrophic forgetting of $D_{f}$. In case of gradient ascent, the model is finetuned using negative of the models gradients on the forget set. As the experiments cover 3 different domains, we use the suitable models in each of the experiments. We use ResNet18~\cite{he2016deep} in computer vision experiments. We use an LSTM model~\cite{hochreiter1997long} with GLOVE embedding~\cite{pennington2014glove} for NLP experiments. We use a Temporal Fusion Transformer (TFT)~\cite{lim2021temporal} for the forecasting experiments. 

\subsection{Experimental Setup}
All the experiments are performed on NVIDIA Tesla-A100 (80GB). The $\lambda$ is set to 50 for Blindspot method in all experiments. The ablation study for different values of $\lambda$ is available in the Supplementary material. We discuss the experimental setup followed in each dataset below.

\textbf{AgeDB and IMDBWiki.} We train the model for 100 epochs with initial learning rate of 0.01 and reduce it on plateau by a factor of 0.1. The models are optimized on L1-loss with Adam optimizer. In FineTune, 5 epochs of training is done with a learning rate of 0.001. We run gradient ascent for 1 epoch with a learning rate of 0.001 on the AgeDB dataset. In Gaussian Amnesiac, 1 epoch of amnesiac learning is done with a learning rate of 0.001. In Blindspot, the blindspot model is trained for 2 epochs with a learning rate of 0.01. Subsequently, 1 epoch of unlearning is performed on the original model with a learning rate of 0.001.\par

\textbf{STS-B SemEval-2017.} The model is trained for 100 epochs with initial learning rate of 0.01 and reduced on plateau by a factor of 0.1. The model is optimized on mean squared error (MSE) loss with Adam optimizer. In FineTune, 10 epochs of training is done with a learning rate of 0.001. In Gaussian Amnesiac, 10 epochs of amnesiac learning is done with a learning rate of 0.001. In Blindspot, the blindspot model is trained for 10 epochs with a learning rate of 0.01 and thereafter, 10 epoch of unlearning is performed on the original model with learning rate of 0.001.\par

\textbf{Electricity Load.} The data points are normalized before training and analysis. We train the model for 10 epochs with initial learning rate of 0.001 and reduce it by 1/10 after every 3 epochs. The model is optimized on Quantile Loss with Adam optimizer. The model predicts 3 quantiles 0.1, 0.5, and 0.9. The data and the model can be used for multi-horizon forecasting but we only forecast for single horizon (one time step) for simplicity. In FineTune, 1 epoch of training is done with a learning rate of $10^{-6}$. In Gaussian Amnesiac, 1 epoch of amnesiac learning is done with a learning rate of $10^{-6}$. In Blindspot method, the blindspot model is trained for 1 epoch with a learning rate of $10^{-5}$ and then 1 epoch of unlearning is performed on the original model with learning rate of $10^{-6}$.

\begin{figure*}[t]
\centering
    \includegraphics[width=0.30\textwidth]{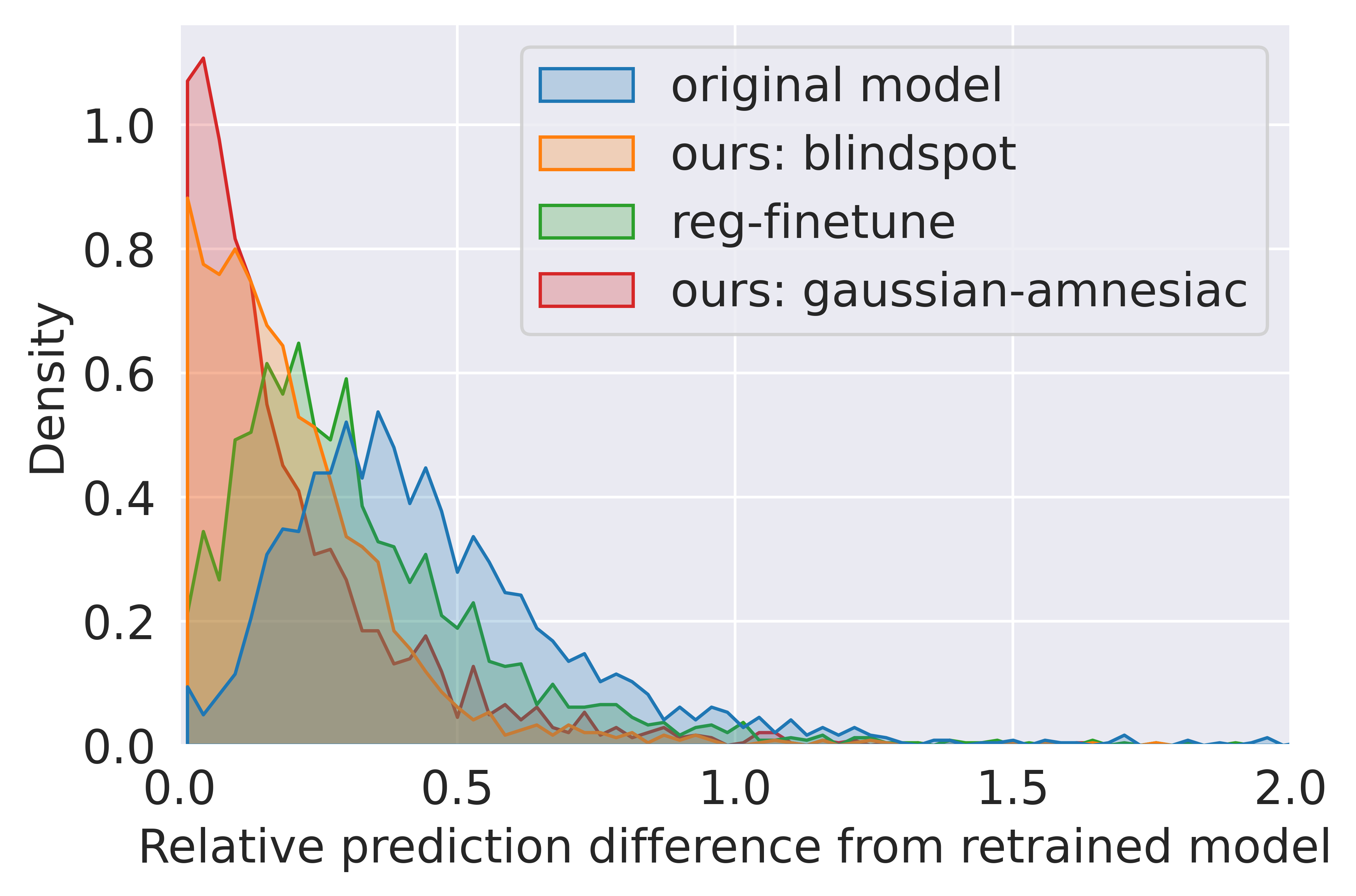}
    \includegraphics[width=0.30\textwidth]{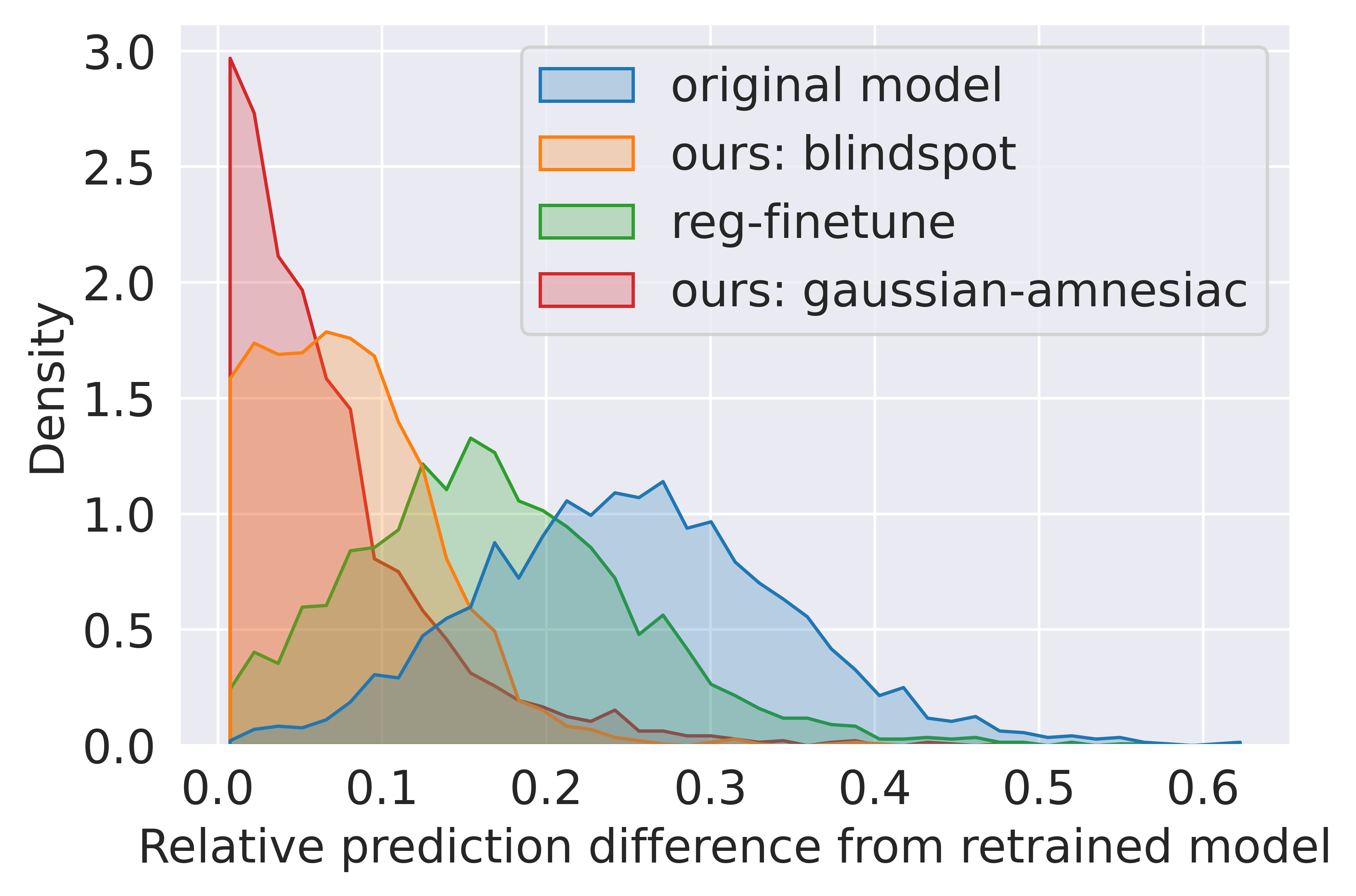}
\caption{Density curves for relative difference between predictions by the unlearning methods and the retrained model on each forget sample in AgeDB. \textit{Density close to 0 represents a better unlearning method.} Left: 0 to 30 age band forgetting, Right: 60 to 100 age band forgetting}
\label{fig:agedbgraphs}
\end{figure*}

\subsection{Results and Analysis}
\textbf{AgeDB and IMDBWiki.} The unlearning result in AgeDB and IMDBWiki dataset for the proposed and baseline methods is presented in~Tables~\ref{tab:agedb} and \ref{tab:imdb-wiki}. In AgeDB, we conduct 3 runs of each experiment and report the standard deviation. Overall, we found the results to be quite stable and conduct single run of all experiments hereafter. We also report the original and retrained model results for comparative analysis. All the three methods obtain similar performance on the retain set ($err\_{D_t^r}$). However, on forget set, the FineTune fares poorly in comparison to the proposed methods. The performance on the forget set ($err\_{D_t^f}$) is ideally expected to be close to the retrained model. The baseline methods FineTune and NegGrad are not able to unlearn properly and report error values much higher than the retrained model. NegGrad may possibly lead to Streisand effect as it has a perfect 0 inference attack probability ($att\_prob$). FineTune has the highest attack probability. In IMDBWiki 0-30 age band unlearning, attack probabilities on FineTune, Gaussian Amnesiac, and Blindspot are 0.26, 0.14, and 0.07, respectively. Whereas, our methods report attack probability closer to the retrained model. While unlearning the age group 60-100 in AgeDB, retrained model's error on forget set is 22.87. The Gaussian Amnesiac and Blindspot unlearning are very close with their respective errors as 21.01 and 25.8. The FineTune method error is 13.5 i.e., it retains most of the initial performance on the forget set. NegGrad's error is 34.87 which is much higher than the retrained model's 22.87 and thus, again suggesting Streisand Effect. The FineTune performs poorly in terms of Wasserstein distance ($w\_dist$) as well. The $w\_dist$ of FineTune is the worst among all the methods (\textit{e.g., 7.40 vs Gaussian Amnesiac's 3.74 and 7.40 vs Blindspot's 1.90 in AgeDB 0-30 band forgetting}).\par
Figure~\ref{fig:agedbgraphs} depicts the density curves for the relative difference between forget data predictions by the unlearning methods and retrained model. The Gaussian Amnesiac has the highest density around 0 in both cases, 0-30 forgetting and 60-100 forgetting. It is followed by the Blindspot method and regular Amnesiac method. Fine-tuning and the original model's curves are farther from 0 and thus suggest a very dissimilar prediction distribution from the retrained model. We discuss the differences between the proposed Gaussian Amnesiac and Regular Amnesiac~\cite{graves2021amnesiac} in Appendix~\ref{sec:ablation_gaussian_vs_regular_Amnesiac}.\par

We also compare the AIN metric for all the unlearning methods. From Table~\ref{tab:imdb-wiki} we can observe that the proposed Blindspot method is quite similar to the retrained model in terms of AIN. Our method has AIN closest to 1 in all the cases in AgeDB 0-30 band forgetting. This is much better in comparison to 0.33 AIN in FineTune and 0.66 AIN in Gaussian Amnesiac.\par

\textbf{STS-B SemEval 2017.}
The unlearning results on STS-B dataset is presented in Table~\ref{tab:sts-b}. Forgetting similarity bands 0-2 induces a significant impact on the forget set error and the proposed Blindspot is the closest in this regard to the retrained model (\textit{Error on forget set: Retrained model: 2.75, Blindspot: 2.47, FineTune: 2.35, Gaussian Amnesiac: 2.30}). Similarly, Blindspot unlearning has the lowest $att\_prob$ on all unlearning cases. For example, while forgetting samples from year 2015, $att\_prob$ in Blindspot is 0.34. This is much better in comparison to 0.50 and 0.47 of FineTune and Gaussian Amnesiac, respectively. Our model also have the lowest $w\_dist$ from retrained model on the forget set except in the case of forgetting year 2015 samples. Our Blindspot method also reports significantly better AIN score in comparison to other methods (Table~\ref{tab:sts-b}) which shows that it has achieved high-quality unlearning.

\textbf{Electricity Load.}
Table~\ref{tft-elec} shows unlearning results in electricity load dataset. We unlearn 2 different bands of values from the fully trained model: the first quartile and the last quartile. For both forget sets, the models obtained by Blindspot has the lowest membership attack probability. For example, when forgetting values $>=0.85$, the $att\_prob$ for Blindspot is 0.17 vs 0.36 for Gaussian Amnesiac and 0.29 for Finetune. The performance on forget and retain set in Blindspot is closest to the retrained model while unlearning in the range $<=-0.85$. The quartile loss on the forget set is the higher in FineTune and Gaussian Amnesiac. But this is even higher as compared to the retrained model which may lead to Streisand effect. In this case the $w\_dist$ is also lowest for the Blindspot method i.e., 0.13 vs 2.25 for Gaussian Amnesiac and 0.13 vs 0.54 for Finetune. Unlearning results in the band $>=0.85$ are mixed and no particular method gives the best result in all the metrics. This is due to the presence of a lot of outliers in this data band. The FineTune method gives the loss nearest to the retrained model in retain set and Gaussian Amnesiac is the nearest in terms of loss on forget set. The $w\_dist$ of the Blindspot method is the lowest for the first band ($<=-0.85$) but highest for the second band ($>=0.85$). The AIN score could not be calculated for the forecasting models as the AIN requires the relearning time of the retrained model to come within a specified range of performance of the original model. The retrained models for these were not able to reach the desired performance even after training for very long period.
\section{Conclusion}
We introduce novel unlearning methods for selectively removing information in deep regression models. To the best of the our knowledge, this work presents first such methods for deep regression unlearning. The proposed Blindspot method use a partially trained model along with a copy of the original model to the forget the query samples. The copied model is optimized with three loss functions and the forgetting is induced through the proposed weight optimization process. A Gaussian Amnesiac learning method is also proposed for deep regression unlearning. The experiments and results show that the proposed methods are effective and generalize well to different type of regression problems. Robustness against several privacy attacks were measured to check information leak in the model. Overall, the proposed deep regression unlearning methods show excellent performance on a variety of evaluation metrics measuring the relearning effort, output distribution, and privacy attacks. The proposed methods also outperform the baseline method in four different datasets. The insights on the challenges and the proposed approaches would inspire future works on deep regression unlearning in other applications.

\section*{Acknowledgements}
This research is supported by the National Research Foundation, Singapore under its Strategic Capability Research Centres Funding Initiative. Any opinions,
findings and conclusions or recommendations expressed in
this material are those of the author(s) and do not reflect the views of National Research Foundation, Singapore.
{
    \small
    \bibliography{main}

\begin{thebibliography}{50}
\providecommand{\natexlab}[1]{#1}
\providecommand{\url}[1]{\texttt{#1}}
\expandafter\ifx\csname urlstyle\endcsname\relax
  \providecommand{\doi}[1]{doi: #1}\else
  \providecommand{\doi}{doi: \begingroup \urlstyle{rm}\Url}\fi

\bibitem[Bevan \& Atapour-Abarghouei(2022)Bevan and
  Atapour-Abarghouei]{pmlr-v162-bevan22a}
Bevan, P. and Atapour-Abarghouei, A.
\newblock Skin deep unlearning: Artefact and instrument debiasing in the
  context of melanoma classification.
\newblock In \emph{Proceedings of the 39th International Conference on Machine
  Learning}, volume 162, pp.\  1874--1892. PMLR, 17--23 Jul 2022.

\bibitem[Bishop \& Nasrabadi(2006)Bishop and Nasrabadi]{bishop2006pattern}
Bishop, C.~M. and Nasrabadi, N.~M.
\newblock \emph{Pattern recognition and machine learning}, volume~4.
\newblock Springer, 2006.

\bibitem[Bourtoule et~al.(2021)Bourtoule, Chandrasekaran, Choquette-Choo, Jia,
  Travers, Zhang, Lie, and Papernot]{bourtoule2021machine}
Bourtoule, L., Chandrasekaran, V., Choquette-Choo, C.~A., Jia, H., Travers, A.,
  Zhang, B., Lie, D., and Papernot, N.
\newblock Machine unlearning.
\newblock In \emph{2021 IEEE Symposium on Security and Privacy (SP)}, pp.\
  141--159. IEEE, 2021.

\bibitem[Brophy \& Lowd(2021)Brophy and Lowd]{brophy2021machine}
Brophy, J. and Lowd, D.
\newblock Machine unlearning for random forests.
\newblock In \emph{International Conference on Machine Learning}, pp.\
  1092--1104. PMLR, 2021.

\bibitem[Cao \& Yang(2015)Cao and Yang]{cao2015towards}
Cao, Y. and Yang, J.
\newblock Towards making systems forget with machine unlearning.
\newblock In \emph{2015 IEEE Symposium on Security and Privacy}, pp.\
  463--480. IEEE, 2015.

\bibitem[Carlini et~al.(2022)Carlini, Jagielski, Papernot, Terzis, Tramer, and
  Zhang]{carlini2022privacy}
Carlini, N., Jagielski, M., Papernot, N., Terzis, A., Tramer, F., and Zhang, C.
\newblock The privacy onion effect: Memorization is relative.
\newblock \emph{arXiv preprint arXiv:2206.10469}, 2022.

\bibitem[Cer et~al.(2017)Cer, Diab, Agirre, Lopez-Gazpio, and
  Specia]{cer2017semeval}
Cer, D., Diab, M., Agirre, E., Lopez-Gazpio, I., and Specia, L.
\newblock Semeval-2017 task 1: Semantic textual similarity-multilingual and
  cross-lingual focused evaluation.
\newblock \emph{arXiv preprint arXiv:1708.00055}, 2017.

\bibitem[Chen et~al.(2022{\natexlab{a}})Chen, Sun, Zhang, and
  Ding]{chen2022recommendation}
Chen, C., Sun, F., Zhang, M., and Ding, B.
\newblock Recommendation unlearning.
\newblock In \emph{Proceedings of the ACM Web Conference 2022}, pp.\
  2768--2777, 2022{\natexlab{a}}.

\bibitem[Chen et~al.(2021)Chen, Zhang, Wang, Backes, Humbert, and
  Zhang]{chen2021machine}
Chen, M., Zhang, Z., Wang, T., Backes, M., Humbert, M., and Zhang, Y.
\newblock When machine unlearning jeopardizes privacy.
\newblock In \emph{Proceedings of the 2021 ACM SIGSAC Conference on Computer
  and Communications Security}, pp.\  896--911, 2021.

\bibitem[Chen et~al.(2022{\natexlab{b}})Chen, Zhang, Wang, Backes, Humbert, and
  Zhang]{chen2022graphunlearning}
Chen, M., Zhang, Z., Wang, T., Backes, M., Humbert, M., and Zhang, Y.
\newblock Graph unlearning.
\newblock In \emph{In Proceedings of the ACM SIGSAC Conference on Computer and
  Communications Security (CCS ’22) 2022}, 2022{\natexlab{b}}.

\bibitem[Chundawat et~al.(2023{\natexlab{a}})Chundawat, Tarun, Mandal, and
  Kankanhalli]{chundawat2022can}
Chundawat, V.~S., Tarun, A.~K., Mandal, M., and Kankanhalli, M.
\newblock Can bad teaching induce forgetting? unlearning in deep networks using
  an incompetent teacher.
\newblock In \emph{Proceedings of the AAAI Conference on Artificial
  Intelligence}, 2023{\natexlab{a}}.

\bibitem[Chundawat et~al.(2023{\natexlab{b}})Chundawat, Tarun, Mandal, and
  Kankanhalli]{chundawat2022zero}
Chundawat, V.~S., Tarun, A.~K., Mandal, M., and Kankanhalli, M.
\newblock Zero-shot machine unlearning.
\newblock \emph{IEEE Transactions on Information Forensics and Security},
  2023{\natexlab{b}}.

\bibitem[Fredrikson et~al.(2015)Fredrikson, Jha, and
  Ristenpart]{fredrikson2015model}
Fredrikson, M., Jha, S., and Ristenpart, T.
\newblock Model inversion attacks that exploit confidence information and basic
  countermeasures.
\newblock In \emph{Proceedings of the 22nd ACM SIGSAC conference on computer
  and communications security}, pp.\  1322--1333, 2015.

\bibitem[Ginart et~al.(2019)Ginart, Guan, Valiant, and Zou]{ginart2019making}
Ginart, A., Guan, M.~Y., Valiant, G., and Zou, J.
\newblock Making ai forget you: Data deletion in machine learning.
\newblock In \emph{Advances in neural information processing systems}, pp.\
  3513--3526, 2019.

\bibitem[Golatkar et~al.(2020{\natexlab{a}})Golatkar, Achille, and
  Soatto]{golatkar2020eternal}
Golatkar, A., Achille, A., and Soatto, S.
\newblock Eternal sunshine of the spotless net: Selective forgetting in deep
  networks.
\newblock In \emph{Proceedings of the IEEE/CVF Conference on Computer Vision
  and Pattern Recognition}, pp.\  9304--9312, 2020{\natexlab{a}}.

\bibitem[Golatkar et~al.(2020{\natexlab{b}})Golatkar, Achille, and
  Soatto]{golatkar2020forgetting}
Golatkar, A., Achille, A., and Soatto, S.
\newblock Forgetting outside the box: Scrubbing deep networks of information
  accessible from input-output observations.
\newblock In \emph{European Conference on Computer Vision}, pp.\  383--398.
  Springer, 2020{\natexlab{b}}.

\bibitem[Golatkar et~al.(2021)Golatkar, Achille, Ravichandran, Polito, and
  Soatto]{golatkar2021mixed}
Golatkar, A., Achille, A., Ravichandran, A., Polito, M., and Soatto, S.
\newblock Mixed-privacy forgetting in deep networks.
\newblock In \emph{Proceedings of the IEEE/CVF Conference on Computer Vision
  and Pattern Recognition}, pp.\  792--801, 2021.

\bibitem[Goldman(2020)]{goldman2020introduction}
Goldman, E.
\newblock An introduction to the california consumer privacy act (ccpa).
\newblock \emph{Santa Clara Univ. Legal Studies Research Paper}, 2020.

\bibitem[Graves et~al.(2021)Graves, Nagisetty, and Ganesh]{graves2021amnesiac}
Graves, L., Nagisetty, V., and Ganesh, V.
\newblock Amnesiac machine learning.
\newblock In \emph{Proceedings of the AAAI Conference on Artificial
  Intelligence}, volume~35, pp.\  11516--11524, 2021.

\bibitem[Guo et~al.(2020)Guo, Goldstein, Hannun, and Van
  Der~Maaten]{guo2020certified}
Guo, C., Goldstein, T., Hannun, A., and Van Der~Maaten, L.
\newblock Certified data removal from machine learning models.
\newblock In \emph{International Conference on Machine Learning}, pp.\
  3832--3842. PMLR, 2020.

\bibitem[He et~al.(2016)He, Zhang, Ren, and Sun]{he2016deep}
He, K., Zhang, X., Ren, S., and Sun, J.
\newblock Deep residual learning for image recognition.
\newblock In \emph{Proceedings of the IEEE conference on computer vision and
  pattern recognition}, pp.\  770--778, 2016.

\bibitem[Hochreiter \& Schmidhuber(1997)Hochreiter and
  Schmidhuber]{hochreiter1997long}
Hochreiter, S. and Schmidhuber, J.
\newblock Long short-term memory.
\newblock \emph{Neural computation}, 9\penalty0 (8):\penalty0 1735--1780, 1997.

\bibitem[Izzo et~al.(2021)Izzo, Smart, Chaudhuri, and Zou]{izzo2021approximate}
Izzo, Z., Smart, M.~A., Chaudhuri, K., and Zou, J.
\newblock Approximate data deletion from machine learning models.
\newblock In \emph{International Conference on Artificial Intelligence and
  Statistics}, pp.\  2008--2016. PMLR, 2021.

\bibitem[Kantorovich(1960)]{kantorovich1960mathematical}
Kantorovich, L.~V.
\newblock Mathematical methods of organizing and planning production.
\newblock \emph{Management science}, 6\penalty0 (4):\penalty0 366--422, 1960.

\bibitem[Kim et~al.(2019)Kim, Kim, Kim, Kim, and Kim]{kim2019learning}
Kim, B., Kim, H., Kim, K., Kim, S., and Kim, J.
\newblock Learning not to learn: Training deep neural networks with biased
  data.
\newblock In \emph{Proceedings of the IEEE/CVF Conference on Computer Vision
  and Pattern Recognition}, pp.\  9012--9020, 2019.

\bibitem[Li et~al.(2021)Li, Wang, and Cheng]{pmlr-v130-li21a}
Li, Y., Wang, C.-H., and Cheng, G.
\newblock Online forgetting process for linear regression models.
\newblock In \emph{Proceedings of The 24th International Conference on
  Artificial Intelligence and Statistics}, volume 130 of \emph{Proceedings of
  Machine Learning Research}, pp.\  217--225. PMLR, 13--15 Apr 2021.

\bibitem[Lim et~al.(2021)Lim, Ar{\i}k, Loeff, and Pfister]{lim2021temporal}
Lim, B., Ar{\i}k, S.~{\"O}., Loeff, N., and Pfister, T.
\newblock Temporal fusion transformers for interpretable multi-horizon time
  series forecasting.
\newblock \emph{International Journal of Forecasting}, 37\penalty0
  (4):\penalty0 1748--1764, 2021.

\bibitem[Liu et~al.(2021)Liu, Ma, Yang, Liu, Ma, and Ren]{liu2021revfrf}
Liu, Y., Ma, Z., Yang, Y., Liu, X., Ma, J., and Ren, K.
\newblock Revfrf: Enabling cross-domain random forest training with revocable
  federated learning.
\newblock \emph{IEEE Transactions on Dependable and Secure Computing}, 2021.

\bibitem[Liu et~al.(2022)Liu, Xu, Yuan, Wang, and Li]{liuinfocomright2022}
Liu, Y., Xu, L., Yuan, X., Wang, C., and Li, B.
\newblock The right to be forgotten in federated learning: An efficient
  realization with rapid retraining.
\newblock In \emph{IEEE INFOCOM 2022 - IEEE Conference on Computer
  Communications}, pp.\  1749–1758. IEEE Press, 2022.

\bibitem[Mahadevan \& Mathioudakis(2022)Mahadevan and
  Mathioudakis]{mahadevan2021certifiable}
Mahadevan, A. and Mathioudakis, M.
\newblock Certifiable unlearning pipelines for logistic regression: An
  experimental study.
\newblock \emph{Machine Learning and Knowledge Extraction}, 4\penalty0
  (3):\penalty0 591--620, 2022.

\bibitem[Marchant et~al.(2022)Marchant, Rubinstein, and
  Alfeld]{marchant2022hard}
Marchant, N.~G., Rubinstein, B.~I., and Alfeld, S.
\newblock Hard to forget: Poisoning attacks on certified machine unlearning.
\newblock In \emph{Proceedings of the AAAI Conference on Artificial
  Intelligence}, volume~36, pp.\  7691--7700, 2022.

\bibitem[Mehta et~al.(2022)Mehta, Pal, Singh, and Ravi]{mehta2022deep}
Mehta, R., Pal, S., Singh, V., and Ravi, S.~N.
\newblock Deep unlearning via randomized conditionally independent hessians.
\newblock In \emph{Proceedings of the IEEE/CVF Conference on Computer Vision
  and Pattern Recognition}, pp.\  10422--10431, 2022.

\bibitem[Micaelli \& Storkey(2019)Micaelli and Storkey]{micaelli2019zero}
Micaelli, P. and Storkey, A.~J.
\newblock Zero-shot knowledge transfer via adversarial belief matching.
\newblock \emph{Advances in Neural Information Processing Systems},
  32:\penalty0 9551--9561, 2019.

\bibitem[Mirzasoleiman et~al.(2017)Mirzasoleiman, Karbasi, and
  Krause]{mirzasoleiman2017deletion}
Mirzasoleiman, B., Karbasi, A., and Krause, A.
\newblock Deletion-robust submodular maximization: Data summarization with
  “the right to be forgotten”.
\newblock In \emph{International Conference on Machine Learning}, pp.\
  2449--2458. PMLR, 2017.

\bibitem[Moschoglou et~al.(2017)Moschoglou, Papaioannou, Sagonas, Deng, Kotsia,
  and Zafeiriou]{moschoglou2017agedb}
Moschoglou, S., Papaioannou, A., Sagonas, C., Deng, J., Kotsia, I., and
  Zafeiriou, S.
\newblock Agedb: the first manually collected, in-the-wild age database.
\newblock In \emph{Proceedings of the IEEE Conference on Computer Vision and
  Pattern Recognition Workshop}, volume~2, pp.\ ~5, 2017.

\bibitem[Neel et~al.(2021)Neel, Roth, and Sharifi-Malvajerdi]{neel2021descent}
Neel, S., Roth, A., and Sharifi-Malvajerdi, S.
\newblock Descent-to-delete: Gradient-based methods for machine unlearning.
\newblock In \emph{Algorithmic Learning Theory}, pp.\  931--962. PMLR, 2021.

\bibitem[Nguyen et~al.(2020)Nguyen, Low, and Jaillet]{nguyen2020variational}
Nguyen, Q.~P., Low, B. K.~H., and Jaillet, P.
\newblock Variational bayesian unlearning.
\newblock In \emph{Advances in Neural Information Processing Systems},
  volume~33, 2020.

\bibitem[Nguyen et~al.(2022)Nguyen, Huynh, Nguyen, Liew, Yin, and
  Nguyen]{nguyen2022survey}
Nguyen, T.~T., Huynh, T.~T., Nguyen, P.~L., Liew, A. W.-C., Yin, H., and
  Nguyen, Q. V.~H.
\newblock A survey of machine unlearning.
\newblock \emph{arXiv preprint arXiv:2209.02299}, 2022.

\bibitem[Pennington et~al.(2014)Pennington, Socher, and
  Manning]{pennington2014glove}
Pennington, J., Socher, R., and Manning, C.~D.
\newblock Glove: Global vectors for word representation.
\newblock In \emph{Proceedings of the 2014 conference on empirical methods in
  natural language processing (EMNLP)}, pp.\  1532--1543, 2014.

\bibitem[Ramdas et~al.(2017)Ramdas, Garc{\'\i}a~Trillos, and
  Cuturi]{ramdas2017wasserstein}
Ramdas, A., Garc{\'\i}a~Trillos, N., and Cuturi, M.
\newblock On wasserstein two-sample testing and related families of
  nonparametric tests.
\newblock \emph{Entropy}, 19\penalty0 (2):\penalty0 47, 2017.

\bibitem[Rothe et~al.(2015)Rothe, Timofte, and Van~Gool]{rothe2015dex}
Rothe, R., Timofte, R., and Van~Gool, L.
\newblock Dex: Deep expectation of apparent age from a single image.
\newblock In \emph{Proceedings of the IEEE international conference on computer
  vision workshops}, pp.\  10--15, 2015.

\bibitem[Sekhari et~al.(2021)Sekhari, Acharya, Kamath, and
  Suresh]{sekhari2021remember}
Sekhari, A., Acharya, J., Kamath, G., and Suresh, A.~T.
\newblock Remember what you want to forget: Algorithms for machine unlearning.
\newblock \emph{Advances in Neural Information Processing Systems}, 34, 2021.

\bibitem[Tarun et~al.(2023)Tarun, Chundawat, Mandal, and
  Kankanhalli]{tarun2021fast}
Tarun, A.~K., Chundawat, V.~S., Mandal, M., and Kankanhalli, M.
\newblock Fast yet effective machine unlearning.
\newblock \emph{IEEE Transactions on Neural Networks and Learning Systems},
  2023.

\bibitem[Ullah et~al.(2021)Ullah, Mai, Rao, Rossi, and Arora]{ullah2021machine}
Ullah, E., Mai, T., Rao, A., Rossi, R.~A., and Arora, R.
\newblock Machine unlearning via algorithmic stability.
\newblock In \emph{Conference on Learning Theory}, pp.\  4126--4142. PMLR,
  2021.

\bibitem[Voigt \& Von~dem Bussche(2017)Voigt and Von~dem Bussche]{voigt2017eu}
Voigt, P. and Von~dem Bussche, A.
\newblock The eu general data protection regulation (gdpr).
\newblock \emph{A Practical Guide, 1st Ed., Cham: Springer International
  Publishing}, 2017.

\bibitem[Wang et~al.(2022)Wang, Guo, Xie, and Qi]{wang2022federated}
Wang, J., Guo, S., Xie, X., and Qi, H.
\newblock Federated unlearning via class-discriminative pruning.
\newblock In \emph{Proceedings of the ACM Web Conference 2022}, pp.\  622--632,
  2022.

\bibitem[Warnecke et~al.(2021)Warnecke, Pirch, Wressnegger, and
  Rieck]{warnecke2021machine}
Warnecke, A., Pirch, L., Wressnegger, C., and Rieck, K.
\newblock Machine unlearning of features and labels.
\newblock \emph{arXiv preprint arXiv:2108.11577}, 2021.

\bibitem[Wu et~al.(2022)Wu, Zhu, and Mitra]{wu2022federated}
Wu, C., Zhu, S., and Mitra, P.
\newblock Federated unlearning with knowledge distillation.
\newblock \emph{arXiv preprint arXiv:2201.09441}, 2022.

\bibitem[Ye et~al.(2022)Ye, Yifang, Song, Yang, Liu, Jin, Song, and
  Wang]{ye2022eccvlearning}
Ye, J., Yifang, F., Song, J., Yang, X., Liu, S., Jin, X., Song, M., and Wang,
  X.
\newblock Learning with recoverable forgetting.
\newblock In \emph{Proceedings of the European Conference on Computer Vision},
  2022.

\bibitem[Yu et~al.(2016)Yu, Rao, and Dhillon]{yu2016temporal}
Yu, H.-F., Rao, N., and Dhillon, I.~S.
\newblock Temporal regularized matrix factorization for high-dimensional time
  series prediction.
\newblock In \emph{Advances in neural information processing systems},
  volume~29, 2016.

\end{thebibliography}
    \bibliographystyle{icml2023}
}

\newpage
\appendix
\onecolumn

\appendix
\section{Information Bound for Unlearning using Blindspot Method}
\label{sec_inf_bound}
\begin{figure}[t]
\centering
    \includegraphics[width=0.45\textwidth]{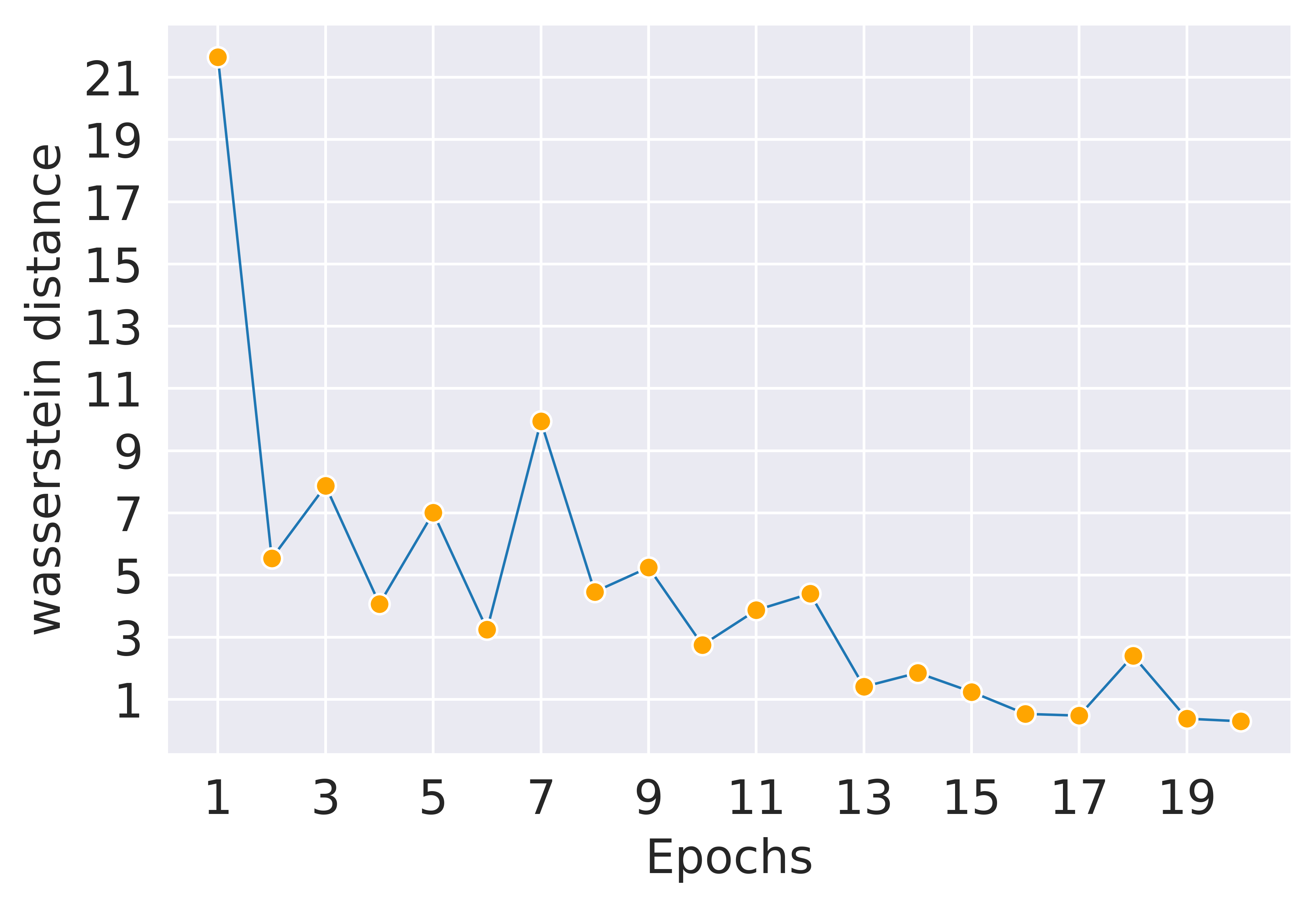}
\caption{Progression of the Wasserstein distance between the predictions of the \textit{Blindspot model} and the \textit{retrained model} on the forget set of range 0-30 in AgeDB.}
\label{fig:wass_prog}
\end{figure}

In Blindspot method, we use a blindspot model to guide the original model with respect to the forget set. The information present in the unlearned model about the forget set after unlearning is bounded by the information present in the blindspot model. Golatkar et al.~\cite{golatkar2020eternal} apply read-out functions and use KL-Divergence between obtained distributions of the unlearned and retrained models on the forget set as a measure of remaining information in classification problems. In our regression setting, we use the identity function as our read-out function i.e., we use the predicted values themselves for distribution comparison. Since KL-Divergence is not applicable until we model a probability distribution function, we use Wasserstein Distance. Let the information present in a model $M$ about a dataset $D$ is denoted by $I(M, D)$. The blindspot model, retrained model, and unlearned model are denoted by $M_b$, $M_r$, $M_u$. Let the forget set is denoted by $D_f$ and $W$ denotes Wasserstein distance between two distributions then
\begin{equation}\label{eq:formal1}
    I(M_u, D_f) \approx I(M_b, D_f)
\end{equation}
\begin{equation}\label{eq:formal2}
    I(M_b, D_f) \propto W(M_b(D_f), M_r(D_f)
\end{equation}
From Eq.~\ref{eq:formal1} and Eq.~\ref{eq:formal2},
\begin{equation}
    I(M_u, D_f) = k W(M_b(D_f), M_r(D_f)
\end{equation}
where $k$ is a constant of proportionality from Eq.~\ref{eq:formal2}. In Figure~\ref{fig:wass_prog}, we plot a graph to show the Wasserstein distance (between blindspot model and retrained model) with respect to the increasing number of epochs. We observe that with increasing epochs, the blindspot model is reaching closer to the prediction distribution of the retrained model on the forget set. We can express $W(M_b(D_f), M_r(D_f)$ as 
\begin{equation}
    W(M_b(Df), M_r(Df)  \le \epsilon
\end{equation}
and, $\epsilon \propto 1/n$

where $n$ denotes the number of epochs for which blindspot model is trained. If we express $\epsilon$  as $\epsilon = c/n$ then
\begin{equation}
    I(M_u, D_f) \le kc/n
\end{equation}

The amount of information the Blindspot method reveals is bounded by $kc/n$. This implies, more the blindspot model is trained, less information is remaining about the forget set in the model. In our experiments we train the retrained models for 100 epochs, and show that training the blindspot model for as less as 2 epochs yields very good quality unlearning.

\section{Evaluation on Additional Privacy Attacks} 
\subsection{Inversion Attacks} 
\label{sec:inversion}
\begin{figure}[t]
\centering
    \includegraphics[width=0.55\textwidth]{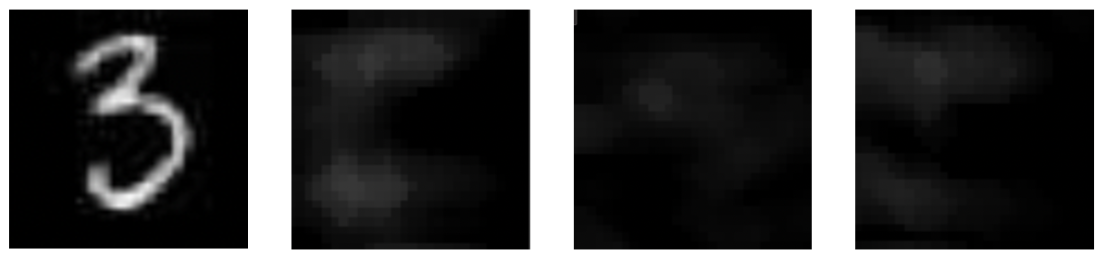}
\caption{From left to right: a sample image, inverted image from the original model, inverted image from the retrained model, inverted image from the unlearned model}
\label{fig:inversion3}
\end{figure}

As the datasets used in the main experiments in the paper do not have specific patterns that can be visually depicted. We conduct regression experiments on MNIST dataset to evaluate robustness of our method to model inversion attacks. We train an AllCNN model which attains a final mean absolute error of 0.08. We evaluate unlearning after forgetting class \textit{3} and \textit{5}. The inversion attack results and comparison are presented in Figure~\ref{fig:inversion3}. The first image is a sample image from class \textit{3} from the MNIST dataset. The second image is an inverted image from the fully trained model. It clearly captures the two edges of \textit{3} which the model is probably using to recognise the shape of \textit{3}. The next two images are of the retrained model and the unlearned model, respectively. These two images do not contain any recognizable pattern.\par

\begin{figure}[t]
\centering
    \includegraphics[width=0.55\textwidth]{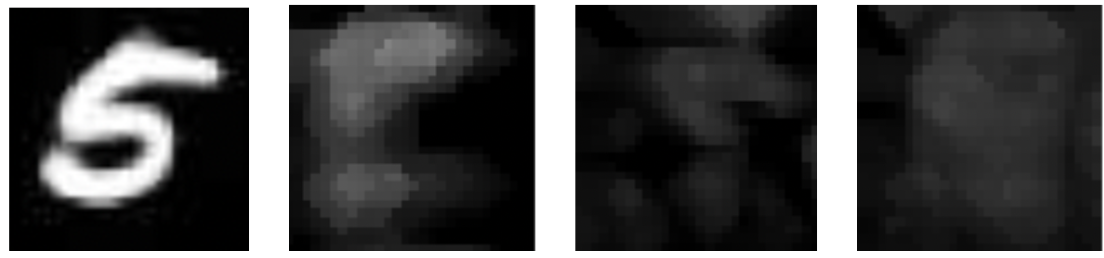}
\caption{From left to right: a sample input, inverted image from the original model, inverted image from the retrained model, inverted image from the unlearned model}
\label{fig:inversion5}
\end{figure}

Another instance of an inversion attack and results are presented in Figure~\ref{fig:inversion5}. While unlearning the shape of \textit{5}, we can see the image obtained after inverting from the fully trained model (second image) has a very recognizable pattern. Whereas, the third and fourth images corresponding to the retrained model and the model obtained from Blindspot unlearning do not have any identifiable patterns. This shows that our method is robust to model inversion attacks.

\subsection{Backdoor Attacks}
We conduct regression experiment on MNIST to observe how the poisoned samples impact the unlearning performance. We add a 4x4 white patch in the bottom right corner on randomly selected 100 images from all classes except class 1 and assign the label 1 to all the patched images. This acts as a backdoor trigger. We then measure the accuracy of the backdoor attack on the model i.e., how many images with patches are predicted with label 1. The higher the accuracy, the more effective is the attack. We unlearn all the images with patches. Besides, we also retrain a model from scratch without the patched images. We compare these two models to observe the attack accuracy.\par

The original model trained with the poisoned samples has an attack accuracy of $98\%$. The model trained without these samples has an attack accuracy of $0\%$. If we use Blindspot method to unlearn the poisoned samples from the original model, the attack accuracy goes down to $0.33\%$. Thus, our unlearning method is successfully able to mitigate the backdoor attack issue when unlearning poisoned samples.
\begin{figure}[t]
\centering
    \includegraphics[width=0.6\textwidth]{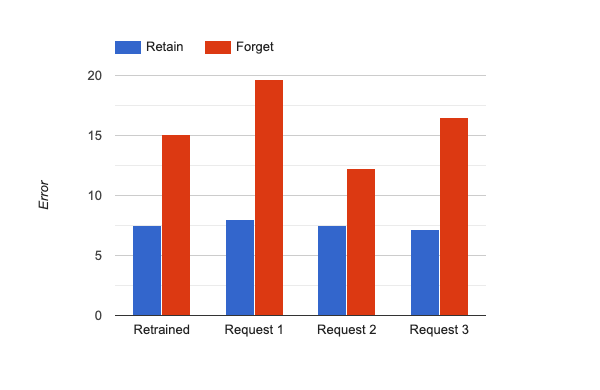}
\caption{Unlearning performance after repeated unlearning requests on AgeDB. Request-1 is to forget age band 0-10, Request-2 is to forget age band 10-20, and Request-3 is to forget age band 20-30. The retrained model is trained from scratch without the 0-30 age band. Our model maintains the retain set error even after Request-3 and the error is similar to the retrained model.}
\label{fig:sequential}
\end{figure}

\begin{figure}[t]
\centering
    \includegraphics[width=0.4\textwidth]{fig/Distribution_Comparison_agedb_0to30.png}
    \includegraphics[width=0.4\textwidth]{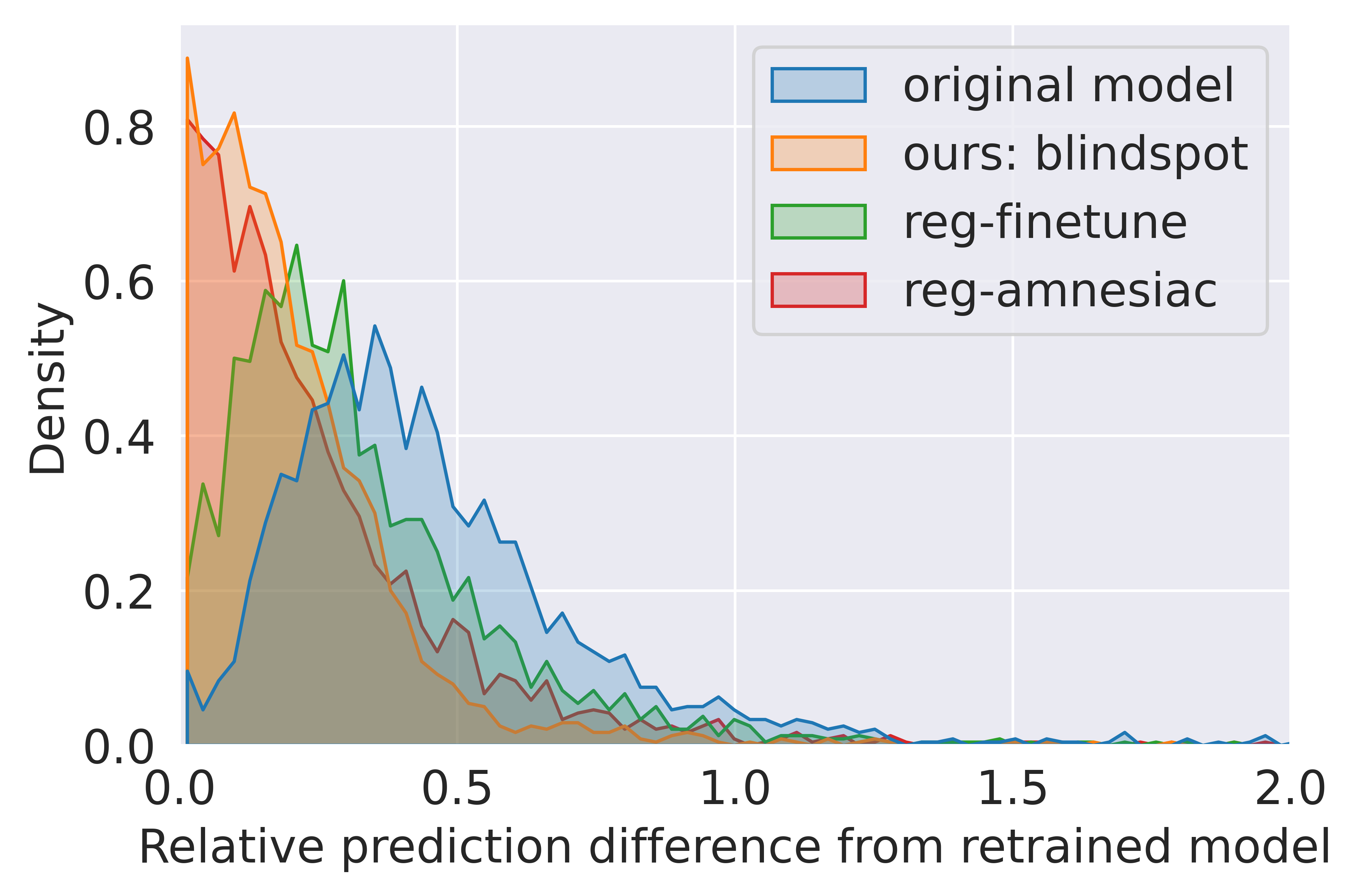}
\caption{Difference between Reg Amnesiac (right) and Gaussian Amnesiac (left) distribution comparison on forget set with retrained model.}
\label{fig:agedb_ablation}
\end{figure}
\section{Sequential Unlearning Requests}
A model might receive multiple unlearning requests at different points in time. Therefore a good unlearning method should perform robustly in case of multiple sequential unlearning requests. Such repeated unlearning should not cause excessive damage to the performance on the retain set. Otherwise, the model might become unusable over the period of time. Figure~\ref{fig:sequential} shows how the proposed Blindspot method handles sequential unlearning requests. The experiment is conducted on AgeDB where first request is to forget 0-10 age band, second request is to forget 10-20 age band, and the third request is to forget 20-30 age band. The retrained model is the one trained from scratch without the 0-30 age band. As clearly visible in Figure~\ref{fig:sequential}, Blindspot unlearning maintains the retain set accuracy after each unlearning request. The result is very much comparable to the retrained model after request 3. This shows the viability of our method for unlearning in continual learning systems.

\section{Gaussian Amnesiac Vs Regular Amnesiac for Regression Unlearning} 
\label{sec:ablation_gaussian_vs_regular_Amnesiac}
As discussed in Section~\ref{sec:reg-amnesiac}, we replace the labels of the samples in the forget set with incorrect ones. In~\cite{graves2021amnesiac}, the incorrect class is randomly sampled from a set of classes other than the correct one. We replicate this for a regression task in Reg Amnesiac by replacing the labels from a uniform distribution of discrete values between [1,101]. In our experiment, we unlearn the band [0,30] in AgeDB. In Gaussian Amnesiac, we use a normal distribution with mean and standard deviation calculated from the labels of the samples in the dataset. Figure~\ref{fig:agedb_ablation} shows how the prediction difference of the Gaussian Amnesiac unlearned model is closer to the retrained model. This is much better in comparison to the unlearned model obtained by Reg Amnesiac. Gaussian Amnesiac's distribution has the highest density around 0 among all the methods. The attack probability in Gaussian Amnesiac is also much lower at 0.13 vs 0.17 of Reg Amnesiac. The Wasserstein distance of Gaussian Amnesiac is 3.74 as compared to 5.20 of Reg Amnesiac. These results establish the superiority and higher quality unlearning obtained by Gaussian Amnesiac over Reg Amnesiac. Some additional ablation studies are also presented later.

\begin{table}[t]
\tiny
\footnotesize
\centering
\caption{Classification unlearning comparison on CIFAR10+ResNet18. Class-level unlearning is done for simple interpretation of results. Class 0 is unlearned for 1-class unlearning, and classes 1-2 are unlearned for 2-class unlearning.}
\begin{tabular}{c|c|c|cc|c}
\hline
\multirow{2}{*}{Method} & \multirow{2}{*}{\# $\mathcal{Y}_{f}$} & \multirow{2}{*}{Accuracy} & Original  & Retrained & Unlearned \\
{} & {} & {} & Model & Model & {Model}\\
\hline
 & \multirow{2}{*}{1} & $D_r$ $\uparrow$ & 77.86 & 78.32 & 71.06\\
UNSIR  & {} & $D_f$ $\downarrow$ & 81.01 & 0 & 0\\
\cline{3-6}
\cite{tarun2021fast} &  \multirow{2}{*}{2} & $D_r$ $\uparrow$ & 78.00 & 79.15 & 73.61\\
{} & {} & $D_f$ $\downarrow$ & 78.65 & 0 & 0\\
\hline
 & \multirow{2}{*}{1} & $D_r$ $\uparrow$ & 77.86 & 78.32 & 78.21\\
{Amnesiac} & {} & $D_f$ $\downarrow$ & 81.01 & 0 & 0\\
\cline{3-6}
\cite{graves2021amnesiac}&  \multirow{2}{*}{2} & $D_r$ $\uparrow$ & 78.00 & 79.15 & 79.52\\
 & {} & $D_f$ $\downarrow$ & 78.65 & 0 & 0\\
\hline
{Fisher}  & \multirow{2}{*}{1} & $D_r$ $\uparrow$ & 77.86 & 78.32 & 10.85 \\
{Forgetting} & {} & $D_f$ $\downarrow$ & 81.01 & 0 & 0 \\
\cline{3-6}
 \cite{golatkar2020eternal} &   \multirow{2}{*}{2} & $D_r$ $\uparrow$ & 78.00 & 79.15 & 7.98\\
 {} & {} & $D_f$ $\downarrow$ & 78.65 & 0 & 0\\
 \hline
 \hline
 & \multirow{2}{*}{1} & $D_r$ $\uparrow$ & 77.86 & 78.32 & 77.71\\
  {Blindspot}& {} & $D_f$ $\downarrow$ & 81.01 & 0 & 10.5\\
\cline{3-6}
{Unlearning}&  \multirow{2}{*}{2} & $D_r$ $\uparrow$ & 78.00 & 79.15 & 80\\
{(\textbf{ours})} & {} & $D_f$ $\downarrow$ & 78.65 & 0 & 12.12\\
\hline
\end{tabular}
\label{Class-UnLearning}
\end{table}

\begin{figure*}[t]
\centering
    \includegraphics[width=0.35\textwidth]{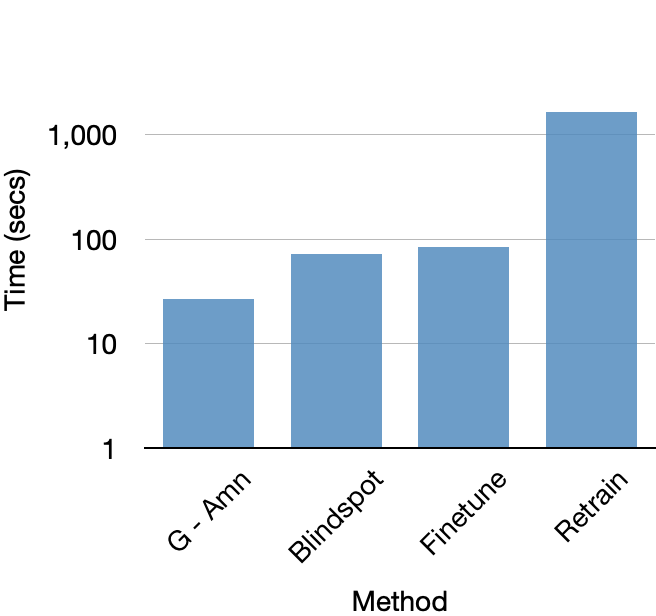}
    \includegraphics[width=0.35\textwidth]{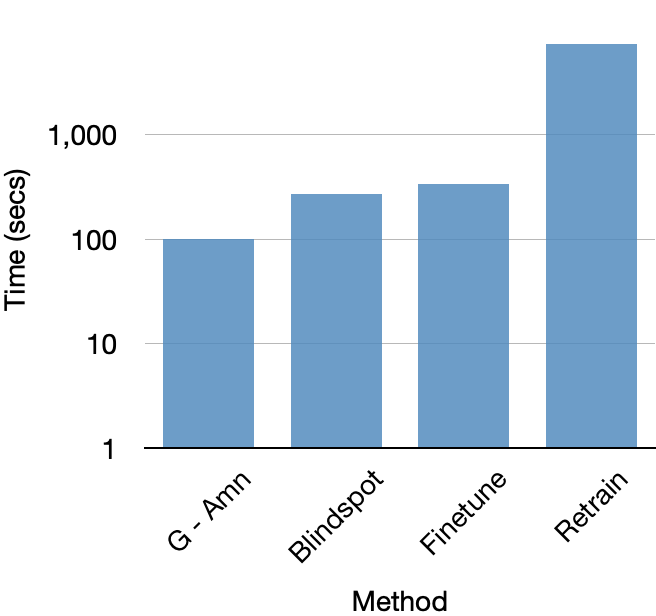}
\caption{Time comparison of different unleaning methods on AgeDB (left) and IMDBWiki (right). Forgetting 0-30 band in  ResNet18. G-Amn: Gaussian Amnesiac. The Y-axis is in logarithmic scale in this figure).}
\label{fig:time_comparison}
\end{figure*}

\section{Viability of Blindspot in Classification Unlearning Tasks}
We perform class-level unlearning with the Blindspot method and show the results in Table~\ref{Class-UnLearning}. We compare the result with existing classification unlearning methods~\cite{tarun2021fast,graves2021amnesiac,golatkar2020eternal}. The forget set accuracy in Blindspot is quite high in comparison with the existing methods. 
The $D_{f}$ accuracy in 1-class and 2-class unlearning is 10.5\% and 12.12\%, respectively. These values should be closer to zero. It appears there is scope to extend the Blindspot method and make it effective for classification unlearning as well. This could be a future scope of this work. 

\section{Efficiency Analysis}
In all the experiments, we use NVIDIA Tesla A100, 80GB GPU. The training time comparison between different unlearning methods are shown in~Figure~\ref{fig:time_comparison}. The training run-time is computed for ResNet18 on AgeDB and IMDBWiki. The original training and retraining is done for 100 epochs. The blindspot model is trained for 2 epochs and the unlearning step is run for 1 epoch. In AgeDB, retraining takes 1666 seconds, fine tuning requires 84 seconds, Gaussian Amnesiac requires 27 seconds, and the Blindspot method requires 72 seconds. The proposed Blindspot method is $>20\times$ faster than retraining and Gaussian Amnesiac is $>60\times$ times faster than retraining. Similarly, on IMDBWiki, the runtime is at least $>20\times$ faster in both Gaussian Amnesiac and Blindspot. The Gaussian Amnesiac is the most efficient in both cases, followed by Blindspot unlearning method. Note that in Figure~\ref{fig:time_comparison} the Y-axis (time) is in logarithmic scale.

\section{Additional Ablation Study}
\subsection{Effect of using different values of $\lambda$} 
We show Blindspot results with a variety of $\lambda$ values in Table~\ref{table:lambda_ablation} and Table~\ref{table:lambda_ablation2}. The results are perceptually bad only in cases of either very low values (0 and 5) or very high values ($\geq75$) of $\lambda$. For other values of $\lambda$ (10-50), the change in the $\lambda$ value does not drastically influence the performance. 

\subsection{Effect of using different \% of retain data in blindspot model}
Table \ref{tab:retain_data_ablation} shows the experimental results on varying amount of retain data  for AgeDB 0-30 forgetting. The amount of retain data seems to be directly proportional to the quality of unlearning and Streisand Effect. Most Streisand Effect is observed when no retain data (0\% $D_r$) is used and this can be seen through Wasserstein Distance with the retrained model and AIN. Wasserstein Distance decreases with an increase in the amount of retain data used. Membership Attack probabilities are 0.0, and thus showcase the Streisand effect for all cases except 100\% $D_r$.

\begin{table}[t]
\footnotesize
\centering
\caption{Effect of using different values of $\lambda$ for 0 to 30 age band unlearning in AgeDB. The results are reported after a single run.}
\begin{tabular}{c|c|c|c|c|c}
\hline
$\lambda$ & $err\_{D_t^r}$  & $err\_{D_t^f}$  & $att\_prob$ & $w\_dist$  & AIN\\
\hline
{0} & {7.53} & {14.73} & {0.11} & {3.16} & {0.33} \\
\hline
{5} & {7.38} & {16.74} & {0.12} & {2.91} & {0.33} \\
\hline
{10} & {7.37} & {17.23} & {0.08} & {2.89} & {0.67} \\
\hline
{25} & {7.55} & {17.81} & {0.10} & {2.47} & {0.33} \\
\hline
{50} & {7.63} & {18.27} & {0.02} & {1.90} & {1} \\
\hline
{75} & {7.67} & {18.34} & {0.03} & {1.96} & {1} \\
\hline
{100} & {7.69} & {17.98} & {0.04} & {2.10} & {1} \\
\hline
{125} & {7.61} & {18.02} & {0.03} & {2.02} & {0.33} \\
\hline
{150} & {7.78} & {16.98} & {0.19} & {3.12} & {1.33} \\
\hline
{175} & {7.81} & {18.13} & {0.02} & {2.73} & {1} \\
\hline
{200} & {7.71} & {18.74} & {0.01} & {2.87} & {0.67} \\
\hline
{250} & {7.59} & {18.06} & {0.05} & {2.12} & {1.33} \\
\hline
\end{tabular}
\label{table:lambda_ablation}
\end{table}

\begin{table}[]
\footnotesize
\centering
\caption{Effect of using different values of $\lambda$ for 0 to 30 age band unlearning in AgeDB. The average $w_{dist}$ after \textit{three} runs are reported.}
\resizebox{\columnwidth}{!}{
\begin{tabular}{c|c|c|c|c|c|c|c|c|c|c|c|c}
    \hline
    $\lambda $ & 0 & 5 &10 &25 &50 &75 &100 &125 &150 &175 &200 &250\\
    \hline
    \multirow{2}{*}{$w_{dist}$} & 2.17  & 1.78  & 1.59  & 1.68  &  1.71  &
    1.81  & 2.09  & 2.07  & 2.51  & 2.33 &
    2.21  & 2.74\\ 
    &$\pm$ 0.21 &$\pm$ 0.01 &$\pm$ 0.02 & $\pm$ 0.11&
    $\pm$ 0.09 &$\pm$ 0.13 &$\pm$ 0.02 &$\pm$ 0.06&
    $\pm$ 0.05 &$\pm$ 0.07 &$\pm$ 0.15 &$\pm$ 0.10\\
    \hline
\end{tabular}
}
\label{table:lambda_ablation2}
\end{table}

\begin{table*}[]
\footnotesize
\centering
\caption{Effect of different proportions of retain data in Blindspot unlearning. The experiments are conducted for AgeDB, 0-30 forgetting.}
\begin{tabular}{c|ccccccc}
\hline
\multirow{2}{*}{Metric} & Original & Retrain & {Blindspot} & {Blindspot} & {Blindspot} & {Blindspot} &{Blindspot}\\
{} & Model & Model & {$0\% of D_r$} & {$10\% of D_r$} & {$25\% of D_r$} & {$50\% of D_r$} & {$100\% of D_r$}\\
\hline
$err\_{D_t^r}$ $\downarrow$ & 7.69 & 7.54 & {9.29} & {7.31} & {7.25} & {7.18} & 7.63\\
$err\_{D_t^f}$ $\uparrow$ & 8.11 & 15.1 & {9.99} & {12.44} & {12.02} & {12.47} & {18.27}\\
{$att\_prob \downarrow$} & {0.72} & {0.07} & {0.0} & {0.0} & {0.0} & {0.0} & {{0.02}}\\
$w\_dist \downarrow$ & {11.39} & {-} & {9.25} & {6.14} & {6.36} & {5.82} & {1.90}\\
AIN $\uparrow$ & {-} & {-} &{6} & {0.33} & {0.33} & {0.33} & {1}\\

\hline
\end{tabular}
\label{tab:retain_data_ablation}
\end{table*}

\subsection{Training with different epochs}
We show the results by varying the number of epochs of training for the blindspot model and the number of overall unlearning epochs in Table~\ref{table:epochs_ablation}. The results are shown for AgeDB, 0 to 30 band forgetting from a ResNet18 model. In Table~\ref{table:epochs_ablation}, we can see that increasing the number of epochs beyond 5 does not lead to any significant advantage when the number of unlearning epochs is fixed at 1. Till epoch 5, we see a steady decrease in Wasserstein distance between the unlearned and the retrained model's prediction distribution. This is because the blindspot model becomes more and more similar to the retrained model. Beyond 5 epochs, there is no significant difference between the predictions of the blindspot model and the retrained model.\par

In another setup, we vary the number of unlearning epochs with a fixed blindspot model training at 2 epochs in Table~\ref{table:epochs_ablation}. We observe an increase in the Wasserstein distance with increasing epochs. This is because the blindspot model is quite far from the retrained model in terms of parameter and prediction distribution as we have only trained it for 2 epochs. With more unlearning epochs, our final model moves closer to the blindspot model. This leads to large error and higher Wasserstein distance which is not desirable. When we fix the number of epochs=5 for blindspot model training (Table~\ref{table:epochs_ablation}), more unlearning epochs lead to better unlearning. This is because the blindspot model's parameters are very close to a retrained model's parameter distribution. More training brings the unlearned model closer to this distribution.

\textbf{Takeaway:} More training of the blindspot model brings it closer to the distribution of a retrained model. Whereas, more unlearning epochs brings the unlearned model closer to the blindspot model on the forget set. There exists a trade-off between the blindspot model training epochs, unlearning epochs and the corresponding unlearning time. We show that even very few epochs yield very good results, but further unlearning can be obtained at the cost of compute time.

\begin{table*}[]
\footnotesize
\centering
\caption{We train the blindspot model for different epochs. We also perform overall unlearning for different epochs on ResNet18+AgeDB (unlearning 0 to 30 age band).}
\begin{tabular}{c|c|c|c|c|c|c}
\hline
Blindspot & Unlearning & \multirow{2}{*}{$err\_{D_t^r}$}  & \multirow{2}{*}{$err\_{D_t^f}$}  & \multirow{2}{*}{$att\_prob$} & \multirow{2}{*}{$w\_dist$}  & \multirow{2}{*}{AIN}\\
 Epochs & Epochs & {} & {} & {} & {} & {}\\
\hline
1 & \multirow{5}{*}{1} & 7.48 & 18.53 & 0.16 & 3.67 & 1.07 \\
2 & {} & 7.63 & 18.27 & 0.02 & 1.90 & 1 \\
5 & {} & 7.41 & 14.38 & 0.65 & 1.19 & 1.03 \\
10 & {} & 7.41 & 18.58 & 0.11 & 3.70 & 0.90 \\
20 & {} & 7.24 & 14.89 & 0.30 & 0.80 & 0.97 \\
50 & {} & 7.38 & 16.8 & 0.24 & 2.08 & 1.23 \\
\hline
\multirow{5}{*}{2} & 2 & 7.36 & 20.32 & 0.08 & 5.38 & 1.37 \\
{} & 5 & 7.33 & 21.75 & 0.005 & 6.62 & 1.27 \\
{} & 10 & 7.34 & 21.89 & 0.002 & 6.75 & 1.77 \\
{} & 20 & 7.62 & 22.74 & 0.01 & 7.59 & 1.93 \\
{} & 50 & 7.68 & 22.33 & 0.005 & 7.18 & 1.83 \\
\hline
\multirow{2}{*}{5} & 2 & 7.25 & 7.25 & 0.39 & 1.17 & 0.87 \\
{} & 5 & 7.43 & 14.71 & 0.34 & 1.11 & 1.10 \\
\hline
\end{tabular}
\label{table:epochs_ablation}
\end{table*}

\subsection{Results on multiple deep models per task}
The main paper contains the results on ResNet18+AgeDB, ResNet18+IMDBWiki, LSTM+STS-B, and TFT (Temporal Fusion Transformer)+Electricity Load. We conduct experiments with additional models per task as follows: AllCNN+AgeDB, MobileNetv3+AgeDB, GRU+STS-B, DNN+STS-B. The unlearning results on AllCNN+AgeDB and MobileNetv3+AgeDB is presented in Table~\ref{tab:allcnn_agedb} and Table~\ref{tab:mobilenet_agedb}, respectively. The results are in line with the obtained results for ResNet18 in the main paper. The Blindspot consistently outperforms the Gaussian Amnesiac method across both AllCNN and MobileNetv3 models on AgeDB dataset.\par 

For STS-B dataset, the results on GRU and DNN are presented in Table~\ref{tab:sts-b}. Similar to the LSTM results, the Blindpsot method gives better results in comparison to Gaussian Amnesiac for GRU and DNN models in text similarity benchmark. 

\subsection{Additional analysis with density curves}
\begin{figure}[]
\centering
    \includegraphics[width=0.4\textwidth]{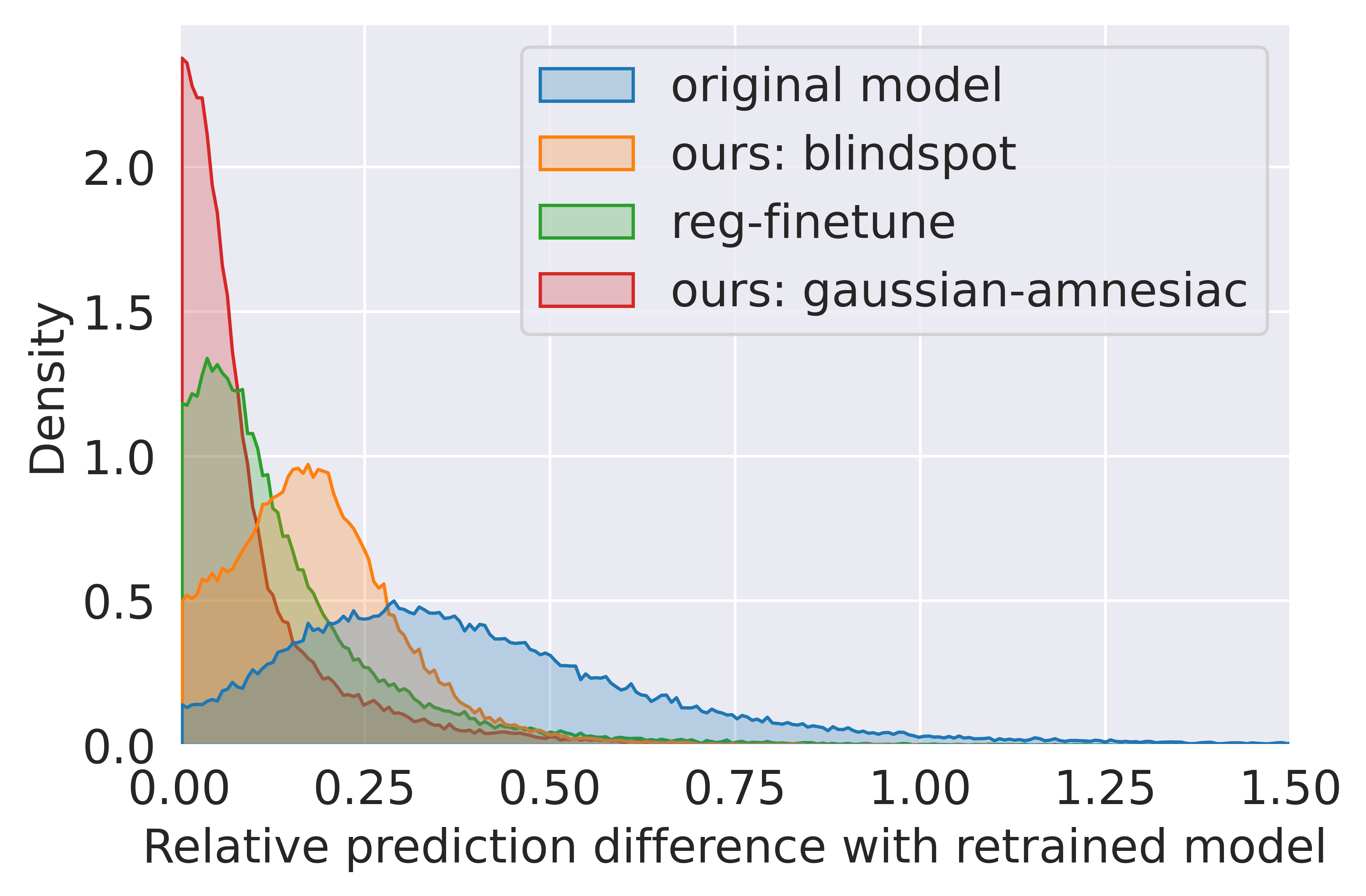}
    \includegraphics[width=0.4\textwidth]{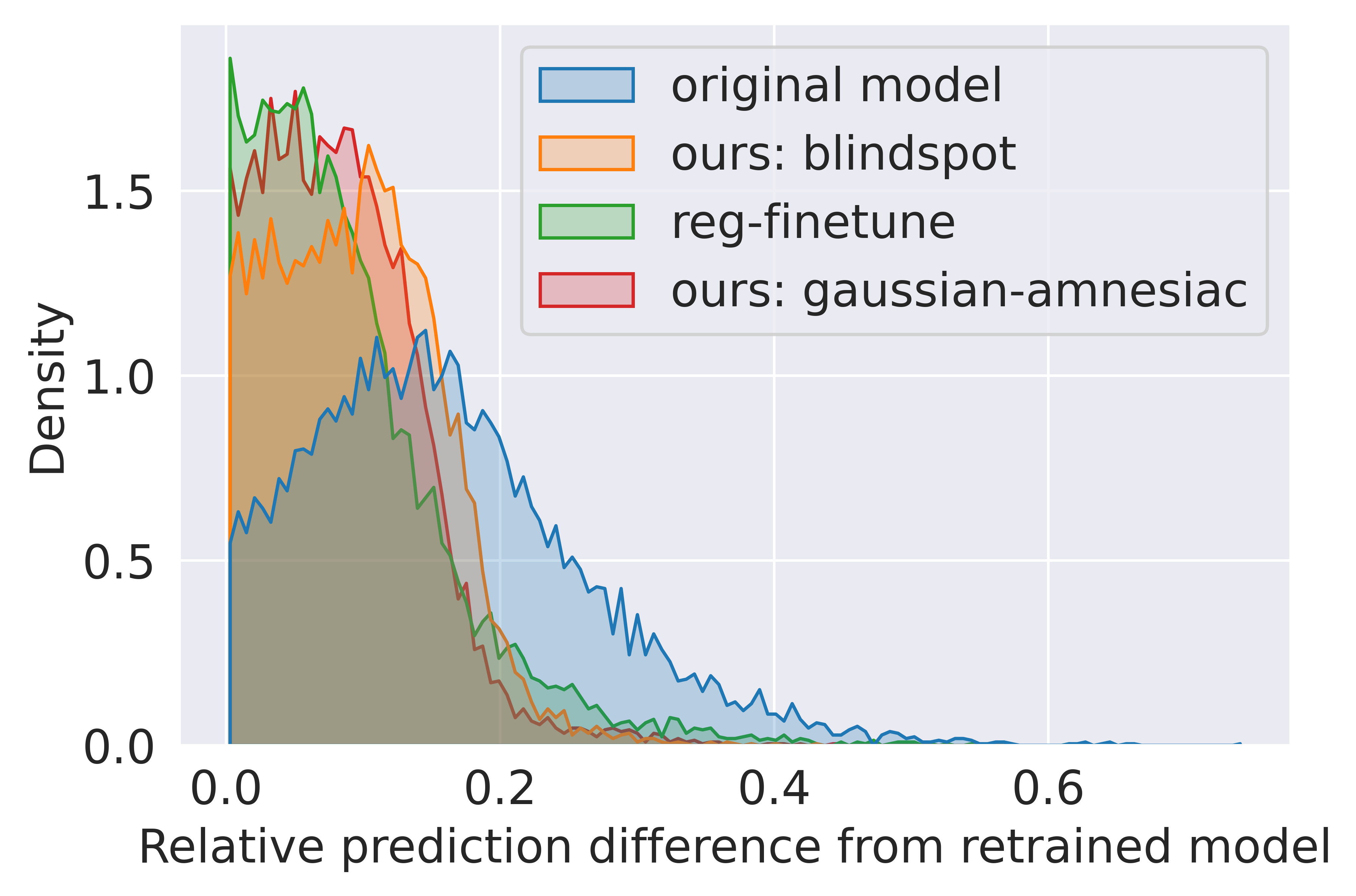}
\caption{Density curves for relative difference between predictions by the unlearning methods and the retrained model on each forget sample in IMDB-Wiki. Left: 0 to 30 age band forgetting, Right: 60 to 100 age band forgetting}
\label{fig:imdbgraphs}
\end{figure}

\begin{figure*}[]
\centering
    \includegraphics[width=0.3\textwidth]{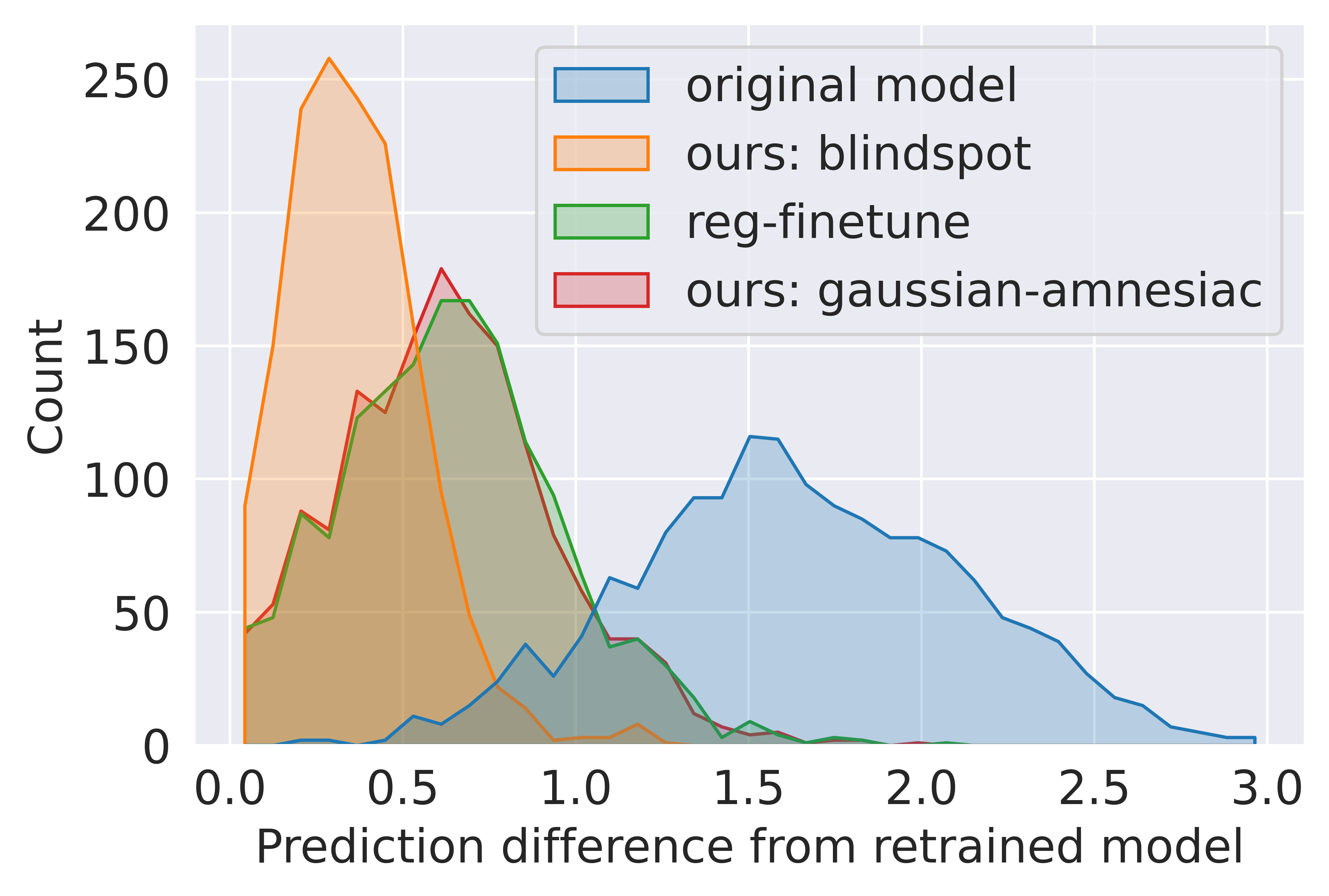}\hspace{-0.5em}
    \includegraphics[width=0.3\textwidth]{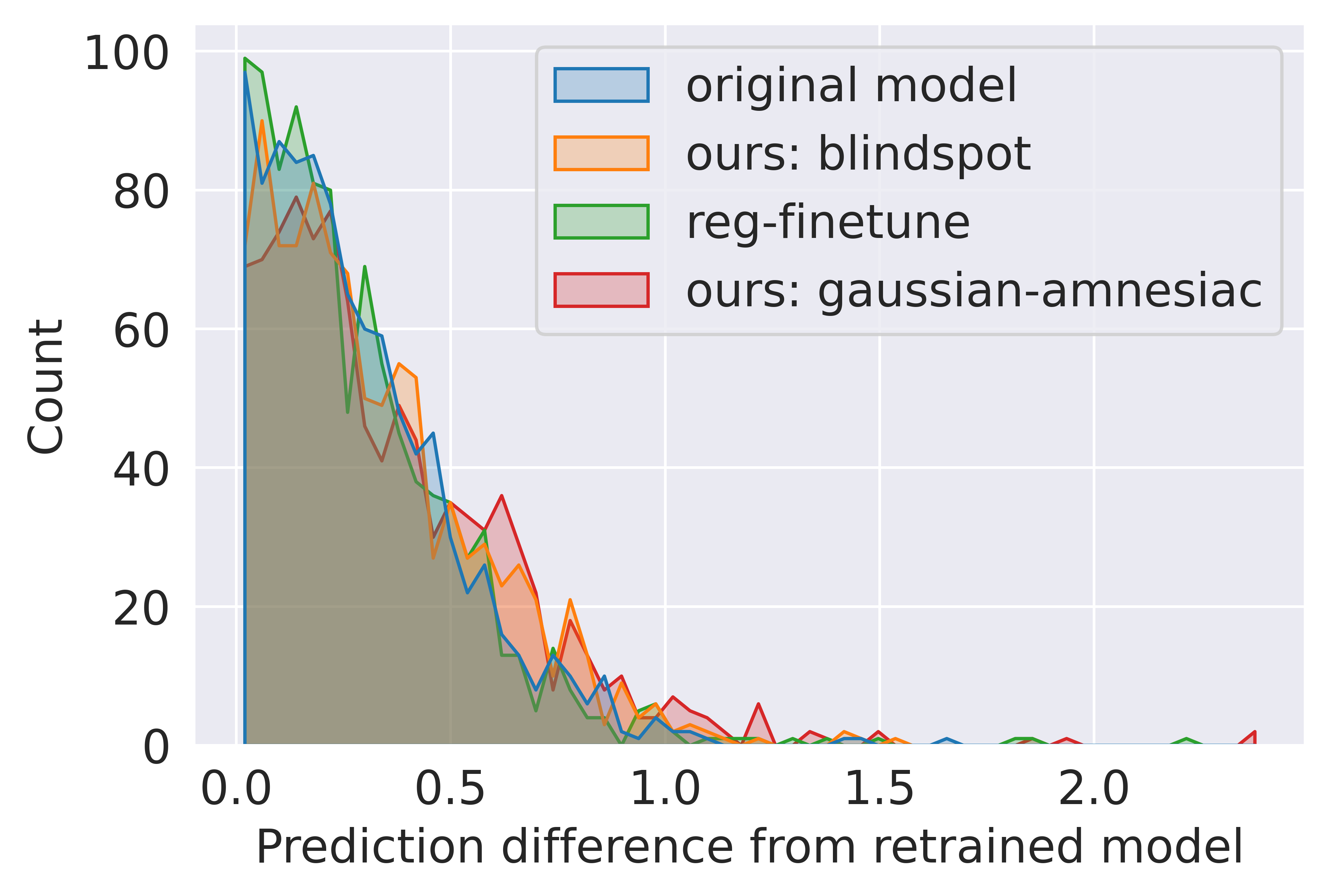}\hspace{-0.5em}
    \includegraphics[width=0.3\textwidth]{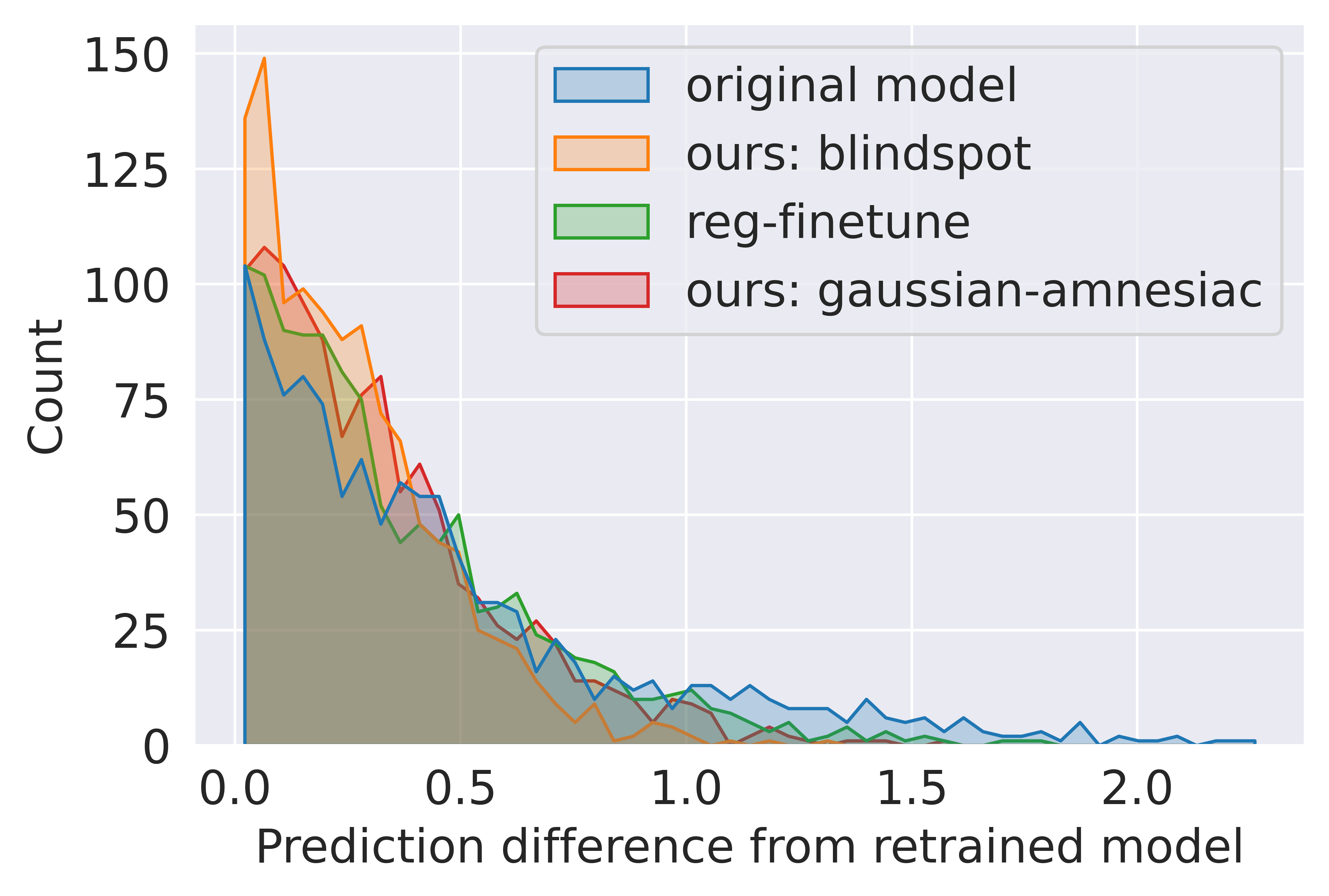}\hspace{-0.5em}
\caption{Distribution plots of differences between predictions by the respective methods and retrained model on each sample of forget set for unlearning in STS-B SemEval-2017 dataset. Left: Band 0-2 forgetting, Middle: 1000 random samples forgetting, Right: Year 2015 samples forgetting}
\label{fig:sts-bgraphs}
\end{figure*}

\begin{figure}[]
\centering
    \includegraphics[width=0.4\textwidth]{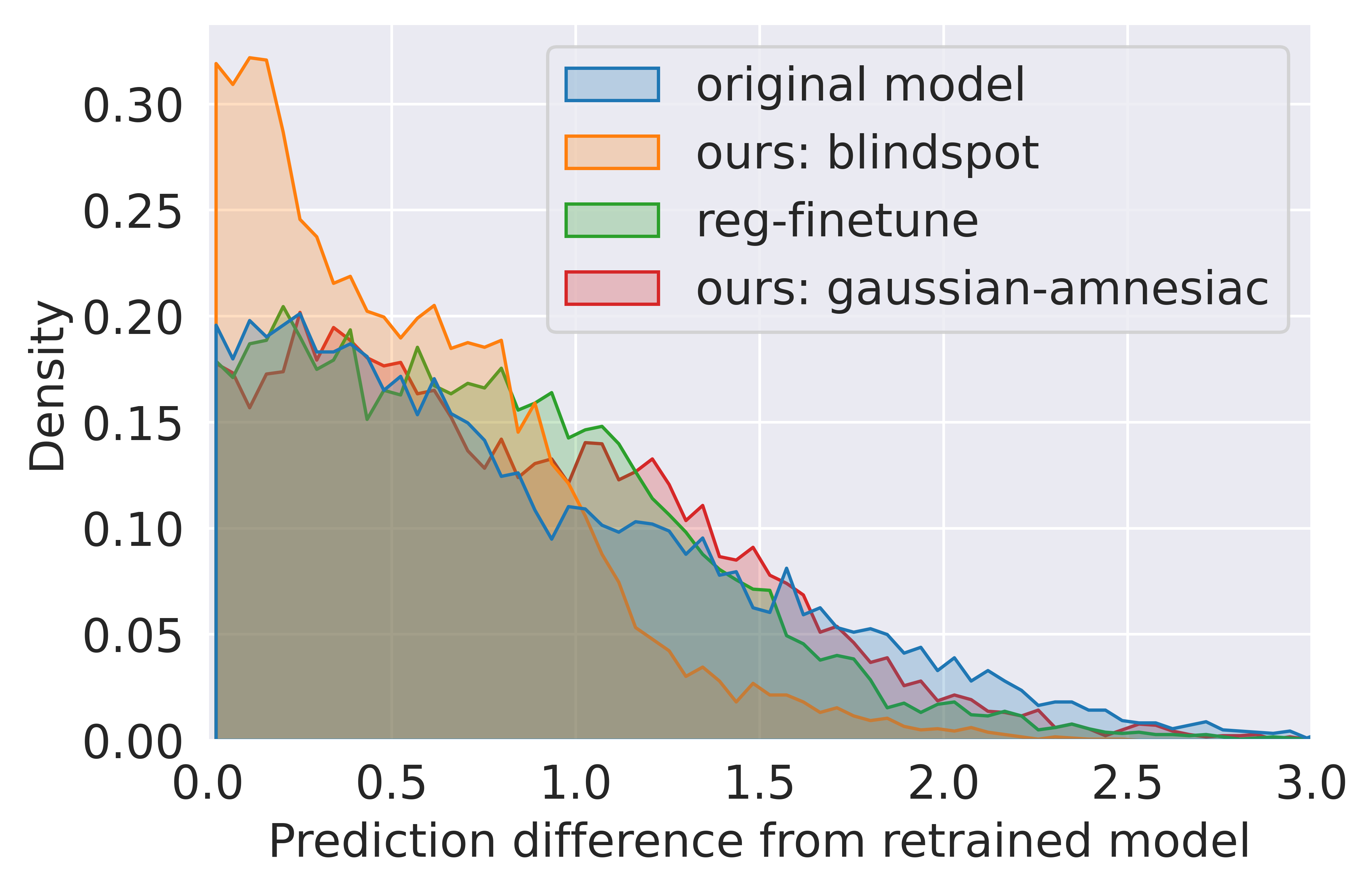}
    \includegraphics[width=0.4\textwidth]{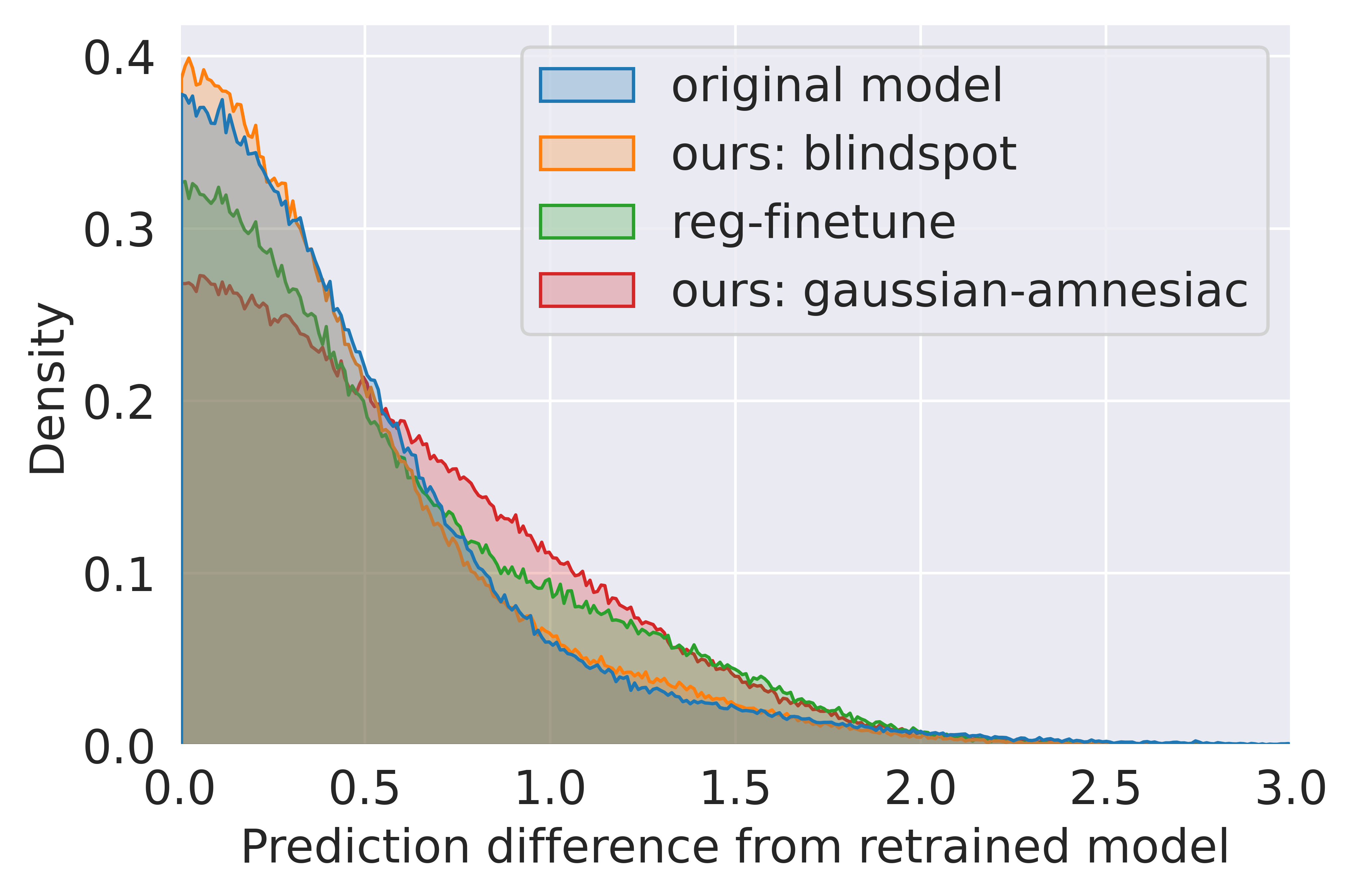}
\caption{Density curves for difference between predictions by the unlearning methods and retrained model on each sample of the forget set for unlearning on UCI Electricity load dataset. Left: forgetting samples with labels $<=-0.85$, Right: forgetting samples with labels $>=0.85$}
\label{fig:tft_dist_comparison}
\end{figure}

 The density curves for difference between predictions by the unlearning methods and retrained model on IMDB-Wiki is shown in Figure~\ref{fig:imdbgraphs}. Original model's curve has the least density around 0. In case of 0-30 forgetting, Gaussian Amnesiac has the highest density around 0, surprisingly followed by finetuning. For 60-100 forgetting, all the methods have very similar density curves. Figure~\ref{fig:sts-bgraphs} depicts the density curves for STS-B SemEval 2017 dataset. We observe that Blindspot has the highest density around 0 i.e., it is the most similar to the retrained model. Only exception is in random sample forgetting where all the models have similar density curves. Figure~\ref{fig:tft_dist_comparison} shows the density curve comparison between all the methods in Electricity Load dataset. The proposed Blindspot method obtains the highest density around 0.

\begin{table*}[]
\footnotesize
\centering
\caption{Unlearning results on AllCNN+AgeDB}
\begin{tabular}{c|c|ccccc}
\hline
Forget &  \multirow{2}{*}{Metric} & Original & Retrain & \multirow{2}{*}{FineTune} & \textbf{Gaussian} &\textbf{Blindspot}\\
{Set} & {} & Model & Model & {} & \textbf{Amnesiac(Ours)} & \textbf{(Ours)}\\

\hline
\multirow{5}{*}{0-30} &  $err\_{D_t^r}$ $\downarrow$ & 9.33 & 11.38 & 9.10  & 10.03 & 10.07\\
{} & $err\_{D_t^f}$ $\uparrow$ & 12.77 & 24.31 & 15.49 & 16.10 & 21.75\\
{} & {$att\_prob \downarrow$} & 0.61 & 0.28 & \textbf{0.25} & 0.33 & \textbf{0.25} \\
{} & $w\_dist \downarrow$ & {5.52} & {-} & 3.58 & 4.54 & \textbf{1.47}\\
{} & AIN $\uparrow$ & {-} & {-} & 1.0 & 1.0 & 1.0\\

\hline
 \multirow{5}{*}{60-100} &  $err\_{D_t^r}$ $\downarrow$ & 9.69 & 10.09 & 8.24  & 8.85 & 9.14\\
{} & $err\_{D_t^f}$ $\uparrow$ & 12.95 & 28.37 & 20.48 & 25.76 & 28.49\\
{} & {$att\_prob \downarrow$} & 0.53 & 0.08 & 0.21 & 0.09 & \textbf{0.06} \\
{} & $w\_dist \downarrow$ & {9.85} & {-} & 4.32 & 1.37 & \textbf{1.09} \\
{} & AIN $\uparrow$ & {-} & {-} & 0.25 & 2.5 & \textbf{1.07}\\
\hline
\end{tabular}
\label{tab:allcnn_agedb}
\end{table*}

\begin{table*}[]
\footnotesize
\centering
\caption{Unlearning results on MobileNetv3+AgeDB}
\begin{tabular}{c|c|ccccc}
\hline
Forget &  \multirow{2}{*}{Metric} & Original & Retrain & \multirow{2}{*}{FineTune} & \textbf{Gaussian} &\textbf{Blindspot}\\
{Set} & {} & Model & Model & {} & \textbf{Amnesiac(Ours)} & \textbf{(Ours)}\\

\hline
\multirow{5}{*}{0-30} &  $err\_{D_t^r}$ $\downarrow$ & 8.56 & 8.00 & 7.35  & 8.82 & 8.63\\
{} & $err\_{D_t^f}$ $\uparrow$ & 9.52 & 17.56 & 16.63 & 14.49 & 19.96\\
{} & {$att\_prob \downarrow$} & 0.67 & 0.28 & 0.34 & 0.17 & \textbf{0.04} \\
{} & $w\_dist \downarrow$ & {4.90} & {-} & \textbf{1.49} & 4.36 & 3.42 \\
{} & AIN $\uparrow$ & {-} & {-} & 0.31 & 1.31 & \textbf{1.15}\\

\hline
 \multirow{5}{*}{60-100} &  $err\_{D_t^r}$ $\downarrow$ & 7.79 & 7.84 & 7.13  & 8.07 & 9.47\\
{} & $err\_{D_t^f}$ $\uparrow$ & 11.58 & 20.68 & 17.92 & 17.20 & 30.40\\
{} & {$att\_prob \downarrow$} & 0.62 & 0.30 & \textbf{0.25} & 0.45 & \textbf{0.28} \\
{} & $w\_dist \downarrow$ & {5.04} & {-} & \textbf{1.46} & 3.63 & 16.44 \\
{} & AIN $\uparrow$ & {-} & {-} & 0.02 & \textbf{0.71} & 0.17\\
\hline
\end{tabular}
\label{tab:mobilenet_agedb}
\end{table*}

\begin{table*}[]
\footnotesize
\centering
\caption{Unlearning results on Semantic Text Similarly Benchmark (STS-B) SemEval-2017 dataset.}
\begin{tabular}{c|c|c|ccccc}
\hline
\multirow{2}{*}{Model} & Forget &  \multirow{2}{*}{Metrics} & Original & Retrain & \multirow{2}{*}{FineTune} & \textbf{Gaussian} &\textbf{Blindspot}\\
{} & {Set} & {} & Model & Model & {} & \textbf{Amnesiac(Ours)} & \textbf{(Ours)}\\
\hline
\multirow{15}{*}{GRU} & \multirow{4}{*}{0-2} & $err\_{D_t^r} \downarrow$ & 2.7 & 0.91 & 1.10 & 1.11 & 1.06\\
{} & {} & $err\_{D_t^f} \uparrow$ & 1.82 & 7.2 & 5.02 & 4.99 & {5.34}\\
{} & {} & {$att\_prob \downarrow$} & 0.52 & 0.004 & 0.05 & 0.04 & \textbf{0.03}\\
{} & {} & $w\_dist \downarrow$  & 1.7 & - & 0.63 & 0.63 & \textbf{0.54}\\
{} & {} & AIN $\uparrow$ & - & - & 1 & 1 & 1\\

\cline{2-8}
{} & {} & $err\_{D_t^r} \downarrow$ & 2.08 & 2.07 & 2.13 & 2.13 & 2.21\\
{} & {Random} & $err\_{D_t^f} \uparrow$ & 1.63 & 2.14 & 1.58 & 1.59 & {1.59}\\
{} & {Samples} & {$att\_prob \downarrow$} & 0.63  & 0.54 & 0.72 & 0.75 & \textbf{0.61}\\
{} & {1000} & $w\_dist \downarrow$  & 0.20 & - & 0.16 & \textbf{0.08} & 0.26\\
{} & {} & AIN $\uparrow$ & - & - & 0.03 & 0.03 & 0.03\\

\cline{2-8}
{} & {} & $err\_{D_t^r} \downarrow$ & 2.08 & 2.15 & 2.10 & 2.15 & 2.17\\
{} & {Year} & $err\_{D_t^f} \uparrow$ & 2.03 & 2.99 & 2.05 & 2.33 & 2.13\\
{} & {2015} & {$att\_prob \downarrow$} & 0.57 & 0.56 & 0.4 & 0.42 & \textbf{0.36}\\
{} & {Samples} & $w\_dist \downarrow$  & 0.39 & - & 0.31 & 0.29 & \textbf{0.27}\\
{} & {} & AIN $\uparrow$ & - & - & 0.03 & \textbf{0.06} & 0.03\\

\hline
\multirow{15}{*}{DNN} & \multirow{4}{*}{0-2} & $err\_{D_t^r} \downarrow$ & 2.79 & 1.02 & 1.19 & 1.37 & 1.17\\
{} & {} & $err\_{D_t^f} \uparrow$ & 1.98 & 7.0 & 5.78 & 4.71 & 5.74\\
{} & {} & {$att\_prob \downarrow$} & 0.54 & 0.01 & 0.05 & 0.05 & \textbf{0.04}\\
{} & {} & $w\_dist \downarrow$  & 1.39 & - & 0.30 & 0.55 & \textbf{0.29}\\
{} & {} & AIN $\uparrow$ & - & - & 0.75 & 0.75 & 0.75\\

\cline{2-8}
{} & {} & $err\_{D_t^r} \downarrow$ & 2.20 & 2.29 & 2.28 & 2.21 & 2.24\\
{} & {Random} & $err\_{D_t^f} \uparrow$ & 2.03 & 2.28 & 2.07 & 2.07 & 2.06\\
{} & {Samples} & {$att\_prob \downarrow$} & 0.63 & 0.49 & 0.67 & 0.58 & \textbf{0.56}\\
{} & {1000} & $w\_dist \downarrow$  & 0.22 & - & 0.11 & 0.13 & \textbf{0.06}\\
{} & {} & AIN $\uparrow$ & - & - & 0.03 & 0.03 & \textbf{0.07}\\

\cline{2-8}
{} & {} & $err\_{D_t^r} \downarrow$ & 2.20 & 2.37 & 2.21 & 2.22 & 2.24\\
{} & {Year} & $err\_{D_t^f} \uparrow$ & 2.56 & 3.58 & 2.90 & 2.90 & 2.97\\
{} & {2015} & {$att\_prob \downarrow$} & 0.46 & 0.36 & 0.44 & 0.46 & \textbf{0.36}\\
{} & {Samples} & $w\_dist \downarrow$  & 0.35 & - & 0.27 & 0.27 & \textbf{0.22}\\
{} & {} & AIN $\uparrow$ & - & - & 1 .0 & 1.0 & 1.0\\

\hline
\end{tabular}
\label{tab:sts-b}
\end{table*}

\end{document}